\PassOptionsToPackage{table}{xcolor}
\documentclass[10pt]{article} 
\usepackage[preprint]{tmlr}

\usepackage{hyperref}
\usepackage{url}

\usepackage[utf8]{inputenc} 
\usepackage[T1]{fontenc}    
\usepackage{hyperref}       
\usepackage{url}            
\usepackage{booktabs}       
\usepackage{amsfonts}       
\usepackage{nicefrac}       
\usepackage{microtype}      
\usepackage{graphicx}
\usepackage{subfigure}
\usepackage{stfloats}
\usepackage{cuted}
\usepackage{capt-of}
\usepackage{xspace}
\usepackage{hyperref}
\usepackage{pgfplots}
\pgfplotsset{compat=1.17}
\usepackage{diagbox} 
\usepackage{amsthm}
\usepackage{amsmath}
\usepackage[capitalize,noabbrev]{cleveref}

\theoremstyle{plain}

\theoremstyle{definition}

\theoremstyle{remark}

\usepackage{soul} 
\usepackage{array}
\usepackage{multirow} 

\definecolor{cellred}{HTML}{6495ED}
\definecolor{cellyellow}{HTML}{B0E0E6}
\definecolor{cellgray}{HTML}{B0C4DE}

\definecolor{lightred}{rgb}{0.39, 0.58, 0.93}
\definecolor{lightyellow}{rgb}{0.69, 0.88, 0.90}
\definecolor{lightgray}{rgb}{0.69, 0.77, 0.87}
\definecolor{mypink}{RGB}{255, 105, 180}

\usepackage{tabularx}
\usepackage{subcaption}
\newcommand{\graycell}[1]{\cellcolor{cellgray}#1}
\newcommand{\redcell}[1]{\cellcolor{cellred}#1}
\newcommand{\yellowcell}[1]{\cellcolor{cellyellow}#1}

\newcommand{\modelname}{Video-3DGS\xspace}

\newcommand{\colmap}{MC-COLMAP\xspace}

\definecolor{blue}{rgb}{0.22, 0.22, 0.95}

\makeatletter
\usepackage{xspace}
\def\@onedot{\ifx\@let@token.\else.\null\fi\xspace}
\DeclareRobustCommand\onedot{\futurelet\@let@token\@onedot}

\hypersetup{
    colorlinks=true,
    urlcolor=mypink, 
}

\def\eg{\emph{e.g}\onedot} 
\def\ie{\emph{i.e}\onedot}

\usepackage[textsize=tiny]{todonotes}

\usepgfplotslibrary{groupplots}

\def\blfootnote{\gdef\@thefnmark{}\@footnotetext}

\title{Enhancing Temporal Consistency in Video Editing by \\Reconstructing Videos with 3D Gaussian Splatting}

\author{\name Inkyu Shin\thanks{Work done during internship at ByteDance.} \email dlsrbgg33@gmail.com \\
      \addr Korea Advanced Institute of Science and Technology
      \AND
      \name Qihang Yu   \email qihang.yu@bytedance.com \\
      \addr ByteDance
      \AND
      \name Xiaohui Shen  \email shenxiaohui@bytedance.com \\
      \addr ByteDance 
      \AND
      \name In So Kweon  \email iskweon77@kaist.ac.kr \\
      \addr Korea Advanced Institute of Science and Technology
      \AND
      \name Kuk-Jin Yoon  \email kjyoon@kaist.ac.kr \\
      \addr Korea Advanced Institute of Science and Technology
      \AND
      \name Liang-Chieh Chen  \email liangchieh.chen@bytedance.com \\
      \addr ByteDance 
      \\}

\def\month{03}  
\def\year{2025} 
\def\openreview{\url{https://openreview.net/forum?id=s1zfBJysbI}} 

\begin{document}

\maketitle

\begin{abstract} 
    Recent advancements in zero-shot video diffusion models have shown promise for text-driven video editing, but challenges remain in achieving high temporal consistency. To address this, we introduce \modelname, a 3D Gaussian Splatting (3DGS)-based video refiner designed to enhance temporal consistency in zero-shot video editors. Our approach utilizes a two-stage 3D Gaussian optimizing process tailored for editing dynamic monocular videos. In the first stage, \modelname employs an improved version of COLMAP, referred to as \colmap, which processes original videos using a Masked and Clipped approach.
    For each video clip, \colmap generates the point clouds for dynamic foreground objects and complex backgrounds. These point clouds are utilized to initialize two sets of 3D Gaussians (Frg-3DGS and Bkg-3DGS) aiming to represent foreground and background views. Both foreground and background views are then merged with a 2D learnable parameter map to reconstruct full views.
    In the second stage, we leverage the reconstruction ability developed in the first stage to 
    impose the temporal constraints on the video diffusion model. This approach ensures the temporal consistency in the edited videos while maintaining high fidelity to the editing text prompt.
    We further propose a recursive and ensembled refinement by revisiting the denoising step and guidance scale used in video diffusion process with \modelname. To demonstrate the efficacy of \modelname on both stages, we conduct extensive experiments across two related tasks: Video Reconstruction and Video Editing.
    \modelname trained with 3k iterations significantly improves video
    reconstruction quality (+3 PSNR, +7 PSNR increase) and training efficiency ($\times1.9$, $\times4.5$ times faster) over  NeRF-based and 3DGS-based state-of-art methods on DAVIS dataset, respectively.
    Moreover, it enhances video editing by ensuring temporal consistency across 58 dynamic monocular videos. Project website is available \href{https://video-3dgs-project.github.io/}{here}.
\vspace{-1mm}
\end{abstract}

\begin{center}
\vspace{-10mm}
\includegraphics[width=0.83\linewidth]{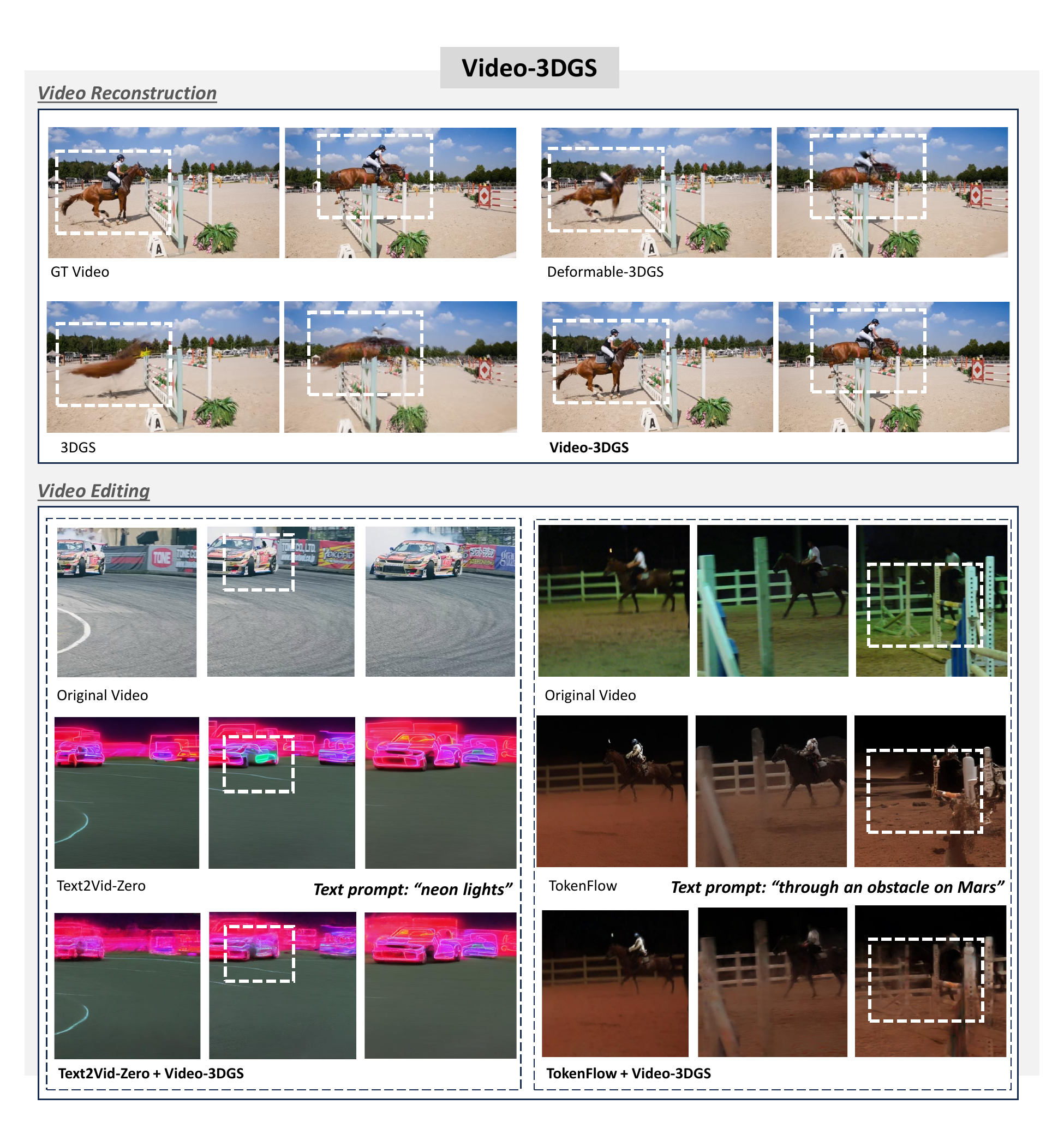}
\vspace{-1mm}
\captionof{figure}{The proposed \modelname expands the capabilities of 3D Gaussian Splatting (3DGS)~\citep{kerbl20233d} to dynamic monocular video scenes, enhancing temporal consistency in both video reconstruction and video editing.
For instance, it consistently captures and reconstructs dynamic objects such as riders and horses (upper section), while also enriching style smoothness in scenarios like drift-car sequences (bottom left) and ensuring structure consistency (bottom right). 
Regions of interest are highlighted by white dashed rectangles.
}
\label{fig:feature-graphic}
\end{center}

\vspace{-1mm}
\section{Introduction}
\label{sec:intro}
\vspace{-1mm}

Advancements in diffusion-based generative models ~\citep{ho2020denoising,rombach2022high} have significantly improved text-driven image editing capabilities~\citep{brooks2022instructpix2pix}. Building on this success, there has been considerable effort to extend these technologies to the video domain~\citep{ho2022video,singer2022make}, which holds practical potential across a broad range of applications, including Film/Entertainment and AR/VR. However, training a video diffusion model from scratch using video datasets is not effective for two main reasons: (i) it lacks a well-constructed video training dataset to handle the diverse distribution of videos in the wild, and (ii) discarding pre-trained image diffusion models would incur a penalty in generating high-quality edited frames for videos. To address the limitations, zero-shot training-free video editing methods~\citep{text2video-zero, tokenflow2023, 2312.04524} that are built on pre-trained image diffusion model have recently been introduced to not only increase the video editing quality but also improve efficiency. Nevertheless, these zero-shot video diffusion models still exhibit low temporal consistency in dynamic videos due to their limited understanding of individual video scenes.
A natural question thus emerges: \textit{Is it possible to design a simple yet effective plug-and-play module to enhance the temporal consistency of each edited video from any zero-shot video editors?}

\begin{figure}[!t]
\vspace{-2mm}
\begin{center}
\includegraphics[width=0.72\linewidth]{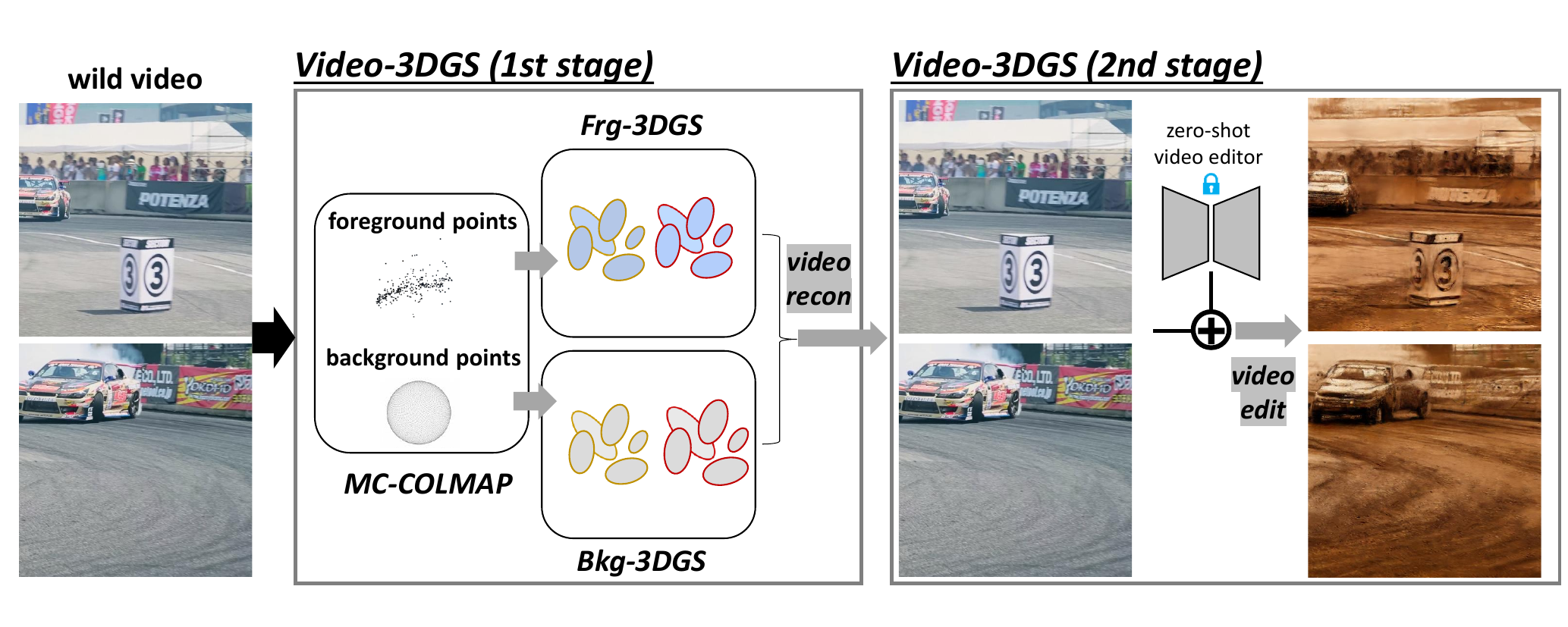}
\caption{
The overall pipeline of \modelname. We aim to design a video-level 3D Gaussian Splatting framework to reconstruct the video scenes (1st stage), which enables high temporal consistency in video editing (2nd stage).
Specifically, \modelname is empowered by the proposed MC-COLMAP that effectively obtains 3D points for foreground moving objects.
The background 3D points are modeled with spherical-shaped random points, surrounding the foreground points.
\modelname utilizes two sets of 3D Gaussians (Frg-3DGS and Bkg-3DGS) to represent foreground and background 3D points, respectively.
A 2D learnable parameter map merges the foreground and background views, rendered from each set of 3D Gaussians.
The merged views enable high-fidelity video reconstruction. Then, we leverage this reconstruction capability into zero-shot video editor to enhance temporal consistency while maintaining high fidelity to text prompt.}
\label{fig:process}
\end{center}
\vspace{-1mm}
\end{figure}

To answer the question, we carefully design \modelname, an innovative approach that leverages per-scene representation power of 3D Gaussian Splatting (3DGS)~\citep{kerbl20233d} to enhance temporal consistency ability for zero-shot video diffusion models. Our approach aims to utilize the structural preservation ability of 3DGS, which is known for explicitly representing and associating multiple views with unified 3D Gaussians. As illustrated in ~\cref{fig:process}, \modelname employs a two-stage 3DGS optimization process to reconstruct dynamic monocular videos and refine the initially edited ones, ensuring enhanced temporal consistency and visual coherence.
\vspace{-1mm}

The first stage begins by addressing the limitations of 3DGS in representing dynamic monocular video scenes, where multiple objects move against complex backgrounds. To solve this issue, new methods adopting 3DGS for dynamic monocular videos are emerging.
Deformable-3DGS~\citep{yang2023deformable3dgs} employed the deformation network to condition 3DGS with time-variable, while 4DGS~\citep{wu20234d} additionally added HexPlane~\citep{cao2023hexplane} encoder before the deformation network.
Despite these attempts to extend 3DGS for monocular video scenes, it is important to acknowledge the persisting limitations, which are summarized as follows.
(i) The fragile dependency on the underlying SfM method, \eg, COLMAP~\citep{schoenberger2016sfm} for deriving 3D points from the entire dynamic videos. 
(ii) The difficulty of accurately representing monocular videos with a single set of 3D Gaussians.
Video frames, particularly those with dynamic motion and intricate background, are infeasible to be represented by a single set of 3D Gaussians, even with the deformation network. As we can observe from ~\cref{fig:feature-graphic}, Deformable-3DGS~\citep{yang2023deformable3dgs} shows unsatisfying video reconstruction results for dynamic moving objects.
Therefore, we first design the framework of \modelname to address the complexities of dynamic monocular video scenes.
One of the key distinctions of the proposed method lies in its ability to excel in ``wilder'' monocular video scenarios (\eg, videos with larger object motion).
To this end, we initially simplify the video sequences with two decomposition strategies. First, spatial decomposition, powered by an off-the-shelf open vocabulary video segmentor, mitigates background clutter~\footnote{background containing complex regions that clutter the foreground objects (\eg, trees in the background).}. Second, temporal decomposition breaks down the entire video sequence into multiple shorter video clips with overlapping frames between neighboring clips.
Given these decompositions, we can effectively extract 3D points of masked foreground moving objects in each video clip by introducing \colmap.
Meanwhile, the cluttered background is modeled with
spherical-shaped random 3D points surrounding
the pre-extracted 3D points of foreground moving
objects.
For each video clip, \modelname utilizes
two sets of 3D Gaussians (Frg-3DGS and Bkg-3DGS) to represent foreground and background
3D points, respectively.
We additionally implement multi-resolution hash encoding~\citep{tiny_cuda_nn} with deformation networks~\citep{yang2023deformable3dgs} on both Frg-3DGS and Bkg-3DGS, which can boost performance and efficiency.
Last but not least, to obtain the final rendered 2D outputs from both 3D Gaussians, we adopt a straightforward and effective 2D learnable parameter map to merge the two rendered images (rendered views from Frg-3DGS and Bkg-3DGS), resulting in faithful video frame representations.

\begin{figure*}[!t]
\begin{center}
\includegraphics[width=0.85\linewidth]{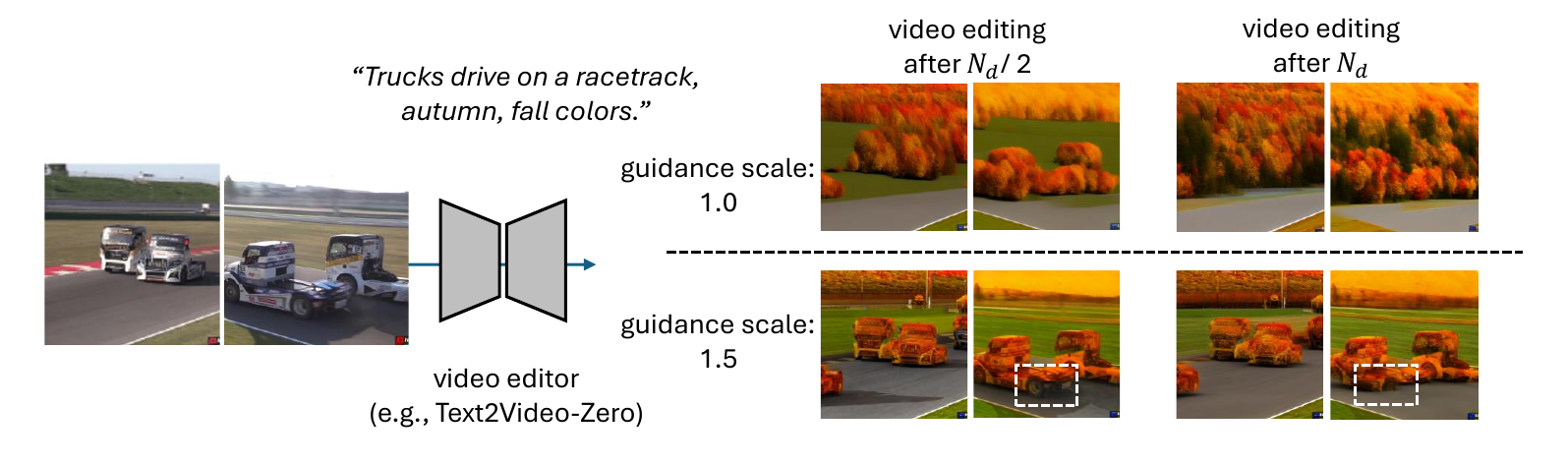}
\caption{
We revisit the key hyperparameters in the video diffusion process: the denoising step ($N_d$) and the guidance scale. Employing a higher denoising step combined with a lower guidance scale (\eg, similar to the image guidance scale in Text2Video-Zero~\citep{text2video-zero}) results in greater fidelity to the editing prompt but compromises structural and temporal consistency, and vice versa. This analysis confirms that the zero-shot video editor model is highly sensitive to these hyperparameters.
}
\label{fig:revisiting}
\end{center}
\end{figure*}

The subsequent stage focuses on seamlessly integrating a pre-optimized \modelname into existing zero-shot video editors. The primary advantage of using a pre-optimized \modelname lies in its ability to apply structured constraints of 3D Gaussians across multiple video frames. More specifically, we maintain the structural components of Frg-3DGS and Bkg-3DGS in a fixed state while selectively fine-tuning the color parameters, such as spherical harmonic coefficients, alongside a 2D learnable parameter map. This fine-tuning process is designed to capture and replicate the style of the initially edited video frames. By applying fixed 3D Gaussian structures with style updates in each clip and smoothing transitions between overlapping frames across neighboring clips, we effectively minimize style flickering throughout the entire video, enhancing temporal consistency.
We have also observed that existing zero-shot video editors exhibit sensitivity to variations in parameters, such as the number of denoising steps and the scale of image or text guidance. These variations can significantly impact the editing outputs, as demonstrated in ~\cref{fig:revisiting}. To optimize effectiveness and reduce parameter sensitivity, we introduce a recursive and ensembled video editing strategy. This involves interspersing \modelname between split denoising steps and updating styles using multiple videos edited under different guidance scales. This further exploration helps to stabilize the editing outcomes across varying parameter settings.

To validate the proposed two-stage \modelname optimization, we tested on corresponding video tasks: video reconstruction and video editing. For video reconstruction, our approach surpasses both NeRF-based and 3DGS-based state-of-the-art methods across 28 DAVIS videos. We then demonstrate the applicability of \modelname's video scene representation to the video editing task in 58 challenging monocular videos sourced from the CVPR 2023 LOVEU Text-Guided Video Editing (TGVE) challenge~\citep{wu2023cvpr}. \modelname consistently enhances editing quality across three off-the-shelf video editors (Text2Video-Zero\citep{text2video-zero}, TokenFlow~\citep{tokenflow2023}, and RAVE~\citep{2312.04524}).



\section{Method}
\label{sec:method}


The meta-architecture of 
\modelname aims to design a video-specific 3DGS, which serves as a plug-and-play refiner for initially edited video from zero-shot video editors. This process is structured into two seamless stages. The first stage of \modelname (\cref{sec:1st}) aims to represent and reconstruct original videos with two integrated components: \colmap (\cref{sec:1st_colmap}) and Foreground/Background 3DGS (\cref{sec:clip3dgs}). 
\colmap plays a pivotal role in generating masked clip-level foreground and background 3D points.
Subsequently, for each clip, two sets of 3D Gaussians, Frg-3DGS and Bkg-3DGS, are initialized and optimized based on these points.
Additionally, a 2D learnable parameter map is employed to merge the foreground and background views rendered from Frg-3DGS and Bkg-3DGS. The resulting views accurately represent the video frames, facilitating the video reconstruction. \cref{sec:ve} details the second stage, where we transform the optimized \modelname into a plug-and-play temporal refiner for video editing. 

\subsection{Preliminary}
\label{sec:pre}
{\bf Diffusion Model}\quad
Diffusion probabilistic models (DPM), introduced by~\citep{sohl2015deep} and further developed in ~\citep{ho2020denoising} represent a class of generative models designed to approximate a data distribution 
$q$ through a progressive denoising process. The model starts with a Gaussian i.i.d noisy image $x_T \sim \mathcal{N}(0, I)$ and the diffusion model $\epsilon_{\theta}$ incrementally denoises this image until it reaches a clean image $x_0$ drawn from the target distribution $q$.
The Deterministic Sampling algorithm (DDIM) described in~\citep{song2020denoising}  initializes the noise process, termed DDIM inversion, by starting from a clean image $x_0$. This approach effectively facilitates image editing through diffusion models. Subsequently, the technique has been adapted for video editing by incorporating temporal constraints. This extension employs several strategies to ensure consistency across frames, including the use of a 3D UNet with cross-frame attention as explored in Text2Video-Zero~\citep{text2video-zero}, diffusion feature propagation as discussed in TokenFlow~\citep{tokenflow2023}, and the grid trick detailed in RAVE~\citep{2312.04524}.

{\bf 3D Gaussian Splatting}\quad
We also present an overview of 3D Gaussian Splatting (3DGS)~\citep{kerbl20233d}.
The method involves the initialization of a set of 3D Gaussians upon 3D point clouds.
The point clouds are derived from Structure from Motion (SfM) techniques, exemplified by COLMAP~\citep{schoenberger2016sfm} with the input $\mathcal{V}$ of a series of $N$ consecutive frames $\mathbf{x}_i$ (\ie, $\mathcal{V}=\{\mathbf{x}_i\}^N_{i\text{=}1}$) in a monocular video sequence.
Then, 3D Gaussians are trained with the function of $G(x, r, s, \sigma, SH)$ containing center position $x$, 3D covariance matrix obtained from quaternion $r$ and scaling $s$, opacity $\sigma$, and spherical harmonic coefficient $SH$.

\subsection{\modelname (1st Stage): Reconstructing Videos with 3DGS}
\label{sec:1st}

\subsubsection{\colmap}
\label{sec:1st_colmap}
We denote the conventional COLMAP function as $R$, which is designed to derive a 3D point cloud $p$, along with the associated camera data $c$ (encompassing both intrinsic ($c^{in}$) and extrinsic ($c^{ex}$) parameters). Furthermore, it outputs a feedback signal $status$, which serves as an indicator of the COLMAP system's success in reconstructing 3D points from the entire video frames $\mathcal{V}$. That is,
\begin{equation}
    p, c, status = R(\mathcal{V})
    \label{eqn:colmap}
\end{equation}
However, the presence of foreground moving objects and cluttered background in video frames poses a significant challenge in extracting consistently matched 3D points across the entire video frames.
This difficulty leads to inaccurate and sparse reconstructions in video outputs, as is evident in our experiments. 
To tackle these challenges, we propose \colmap, a revised version of COLMAP, specifically designed to progressively process video frames.
It effectively minimizes motion and complex background through two key strategies: spatial decomposition and temporal decomposition.

\begin{figure*}[!t]
\begin{center}
\includegraphics[width=0.75\linewidth]{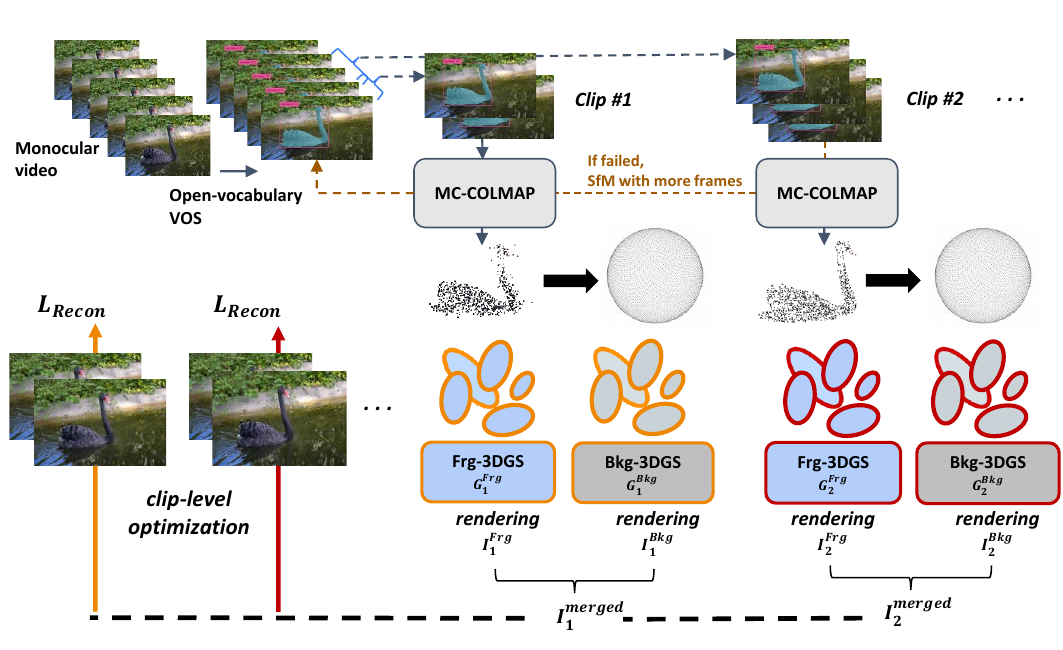}
\caption{
The proposed \modelname (1st stage) comprises two key components. First, to effectively capture 3D point clouds and corresponding frame viewpoints from dynamic monocular videos, we introduce \textbf{M}asked and \textbf{C}lipped COLMAP (MC-COLMAP). This module spatially and temporally decomposes video frames, facilitating the extraction of clip-level foreground points through progressively processing clips. Additionally, we initialize spherical-shaped random background points conditioned on the foreground points.
Second, with these two sets of point clouds, we introduce two distinct sets of 3D Gaussian Splatting (3DGS): Frg-3DGS and Bkg-3DGS, optimized separately for foreground and background points, respectively. Subsequently, we employ a straightforward merging operation to combine the rendered outputs of Frg-3DGS and Bkg-3DGS. We optimize the merged rendered outputs for each clip using the reconstruction loss. 
}
\label{fig:video-3dgs}
\end{center}
\end{figure*}

{\bf Spatial Decomposition}\quad
To reduce the background cluttering effect,
we adopt an off-the-shelf open-vocabulary video object segmentation network $S$ (\eg, DEVA~\citep{cheng2023tracking}) to extract the segmentation masks for foreground moving objects in the video $\mathcal{V}$ as follows:
\begin{equation}
    \mathcal{V}^{f} = S(\mathcal{V}, class)
    \label{eqn:deva}
\end{equation}
where $\mathcal{V}^{f}$ and $class$ denote the extracted segmentation masks of foreground moving objects and the user-guided text prompt (a required input for the segmentation network to specify the target object), respectively. For example, in the top left of~\cref{fig:video-3dgs}, we can consistently extract `Swan' across frames in monocular video using spatial decomposition strategy.


{\bf Temporal Decomposition}\quad
The segmentation masks $\mathcal{V}^{f}$ for foreground objects, which span the entire video sequence, present a processing challenge for COLMAP due to their intricate motion dynamics.
To mitigate this complexity, stemming from diverse motion patterns throughout the video sequence, we partition the video sequence into multiple shorter video clips. This division ensures that objects exhibit reduced motion within each clip, thus facilitating more manageable processing for COLMAP.
Formally, we address this by splitting $\mathcal{V}^{f}$ into multiple $M$ clips $\{\mathcal{V}^f_j\}_{j=1}^M$.

Rather than evenly dividing the video sequence into $M$ clips, each containing $k$ frames, we introduce a progressive scheme to address potential failure cases, such as when the foreground object remains static within a clip, making the $k$ frames inadequate for point cloud extraction or registration in SfM. 
Specifically, we begin with the first clip initialized with the first $k$ frames (\ie, $\mathcal{V}^f_1=\{\mathbf{x}_i\}^k_{i=1}$).
If the COLMAP function $R$ fails to process the clip (\ie, $status_1 \neq$ `Success'), we iteratively include one additional consecutive frame into the current clip until COLMAP returns a `Success' $status$. 
Subsequently, the next clip commences with the last frame of previous clip, also initialized with $k$ frames. This process continues until the entire video sequence is processed.
We term the resulting pipeline as \colmap, which applies COLMAP in a \textbf{M}asked and \textbf{C}lipped manner.
Our \colmap system (denoted as $R_{MC}$) yields multiple sets of masked and clipped 3D point clouds, along with their corresponding views from $M$ clips as demonstrated in the top right of~\cref{fig:video-3dgs} and following equation:
\begin{equation}
    \{{(p_j, c_j)}\}^M_{j=1}=R_{MC}(S(\mathcal{V}, class))
    \label{eqn:rmc}
\end{equation}
We validate the effectiveness of \colmap in~\cref{tab:recon_sum_quan} and~\cref{tab:mccolmap}, demonstrating that \colmap not only improves the success rate over the original COLMAP, but also provides better-initialized 3D points for constructing 3D Gaussians. 

\subsubsection{Foreground and Background 3DGS}
\label{sec:clip3dgs}

{\bf Foreground 3D Gaussians}\quad
For each video clip, the 3D point clouds for foreground moving objects, derived from \cref{eqn:rmc}, serve as  initialization for optimizing foreground 3D Gassuians.
This process yields a set of 3D Gaussians $\{G^{Frg}_j\}^M_{j=1}$, tailored for those foreground point clouds with the corresponding camera views, $\{(p_j, c_j)\}^M_{j=1}$.
With this, we successfully represent the foreground objects using 3D Gaussians, and the subsequent step involves modeling the cluttered background.



{\bf Background 3D Gaussians}\quad
For each video clip, to simply model the cluttered background, we utilize spherical-shaped random point clouds $\{p^{Bkg}_j\}^M_{j=1}$, surrounding the previously extracted foreground 3D points.
These background random point clouds serve as the initialization for optimizing background 3D Gaussians, yielding a set of 3D Gaussians $\{G^{Bkg}_j\}^M_{j=1}$.
Notably, the spherical-shaped random point clouds are defined by two hyper-parameters: number of points $n_{Bkg}$ and its radius $r_i$.
Our method is empirically found robust to those hyper-parameters, and we fix $n_{Bkg}=60k$ points and $r_i$ to 3 times larger than the foreground points' distance that is measured by the maximum Euclidean distance between foreground points.


\textbf{Deformable 3D Gaussians}\quad
Following~\citep{yang2023deformable3dgs}, both the foreground and background 3D Gaussians are enhanced with the deformable network, which effectively leverage the time information, but we extend them for clip-level processing, resulting in:
\begin{equation}
    \begin{split}
    \text{Frg-3DGS: }
    \{G^{Frg}_{j}(\{x_j, r_j, s_j\}+\delta^{Frg}_j, \sigma_j, SH_j)\}^{M}_{j=1} \\ 
    \text{Bkg-3DGS: }
    \{G^{Bkg}_{j}(\{x_j, r_j, s_j\}+\delta^{Bkg}_j, \sigma_j, SH_j)\}^{M}_{j=1}
    \label{eqn:loc_global}
    \end{split}
\end{equation}
where $\delta_j$ is the deformation within the $j$-th clip to transform center $x_j$, rotation $r_j$, and scale $s_j$, according to each clip center and normalized time. The superscript $Frg$ and $Bkg$ denote foreground and background, respectively.
We implement the deformation network with 4D multi-resolution hash encoding~\citep{tiny_cuda_nn}.
Given those two sets of Frg-3DGS and Bkg-3DGS, we reformulate them as Clip-3DGS, where each clip contains its corresponding Frg-3DGS and Bkg-3DGS. That is, $\text{Clip-3DGS}=\{G^{Frg}_j, G^{Bkg}_j\}^M_{j=1}$.
We then process those 3D Gaussians in a clip-by-clip manner.



{\bf Merging Foreground and Background Views with 2D Learnable Parameters}\quad
For the $j$-th clip, we leverage the differentiable point-based rendering technique~\citep{wiles2020synsin}, as outlined in~\citep{kerbl20233d}, to process the two sets of 3D Gaussians $G^{Frg}_j$ and  $G^{Bkg}_j$. 
This enables us to generate two distinct rendered images for each frame within the $j$-th clip.
Specifically, rendering the $i$-th video frame produces two images, $\{I^{Frg}_{i},I^{Bkg}_{i}\}$, derived from  $G^{Frg}_j$ and $G^{Bkg}_j$, respectively.
To seamlessly merge these images, both of dimensions height $H$ and width $W$, we introduce a straightforward yet powerful merging technique. This method employs a 2D learnable parameter $\alpha$ $\in \mathcal{R}^{H\times W}$, facilitating pixel-wise merging with corresponding learnable parameters initialized to a value of 0.5.
Formally, we have: 
\begin{equation}
 I^{merged}_{i} = \alpha_{i}\times I^{Frg}_{i} + (1-\alpha_{i})\times I^{Bkg}_{i}
 \label{eqn:merge}
\end{equation}
where $I^{merged}_{i}$ is the merged result for the $i$-th video frame.

Through the merging operation, we derive merged images for all video frames by optimizing $N$ different $\alpha$ values ($N$ represents the total number of frames in the video).


{\bf Training Losses}\quad
To optimize Frg-3DGS and Bkg-3DGS, we adopt the reconstruction loss, denoted as $L_{recon}$, which comprises two components: $L_{1}$ and $L_{SSIM}$, akin to the approach outlined in~\citep{kerbl20233d}. These components are calculated by comparing three rendered images - the foreground $I^{Frg}$, the background $I^{Bkg}$, and the merged image $I^{merged}$ - against their respective ground truth images.

{\bf Video Reconstruction}\quad
The reconstruction of a video can be achieved by sequentially rendering the Frg-3DGS and Bkg-3DGS for each clip. Subsequently, the rendered images from each frame are merged using a pre-trained alpha parameter (\cref{eqn:merge}).

\subsection{\modelname (2nd Stage): Plug-and-Play Refiner for Editing Videos}
\label{sec:ve}
\textbf{Single-phase Refiner}\quad 
We employ pre-trained 3D Gaussians to preserve the original structure and ensure temporal consistency in video editing. Edited video frames, denoted as 
\begin{equation}
    \{I^{edited}_i\}^N_{i=1}, 
\end{equation}
can be generated by any zero-shot video editor. However, these edits may introduce style inconsistencies or cause original objects to be changed. To address this issue, we fix the positional parameters $(x, r, s)$ and deformation parameters ($\delta$) of both Frg-3DGS and Bkg-3DGS. At the same time, we adjust the color value ($SH$) and opacity ($\sigma$) parameters to match the edited style. This update is achieved by minimizing the reconstruction loss between the rendered images $\{I^{merged}_i\}^N_{i=1}$ and edited frames $\{I^{edited}_i\}^N_{i=1}$ for updating corresponding $SH$ and $\sigma$ as follows:
\begin{equation}
    \min_{\mathbf{SH}, \mathbf{\sigma}}\sum_{i=1}^N L_{\text{recon}}(I^{merged}_i, I^{edited}_i)
    \label{eqn:single_edit}
\end{equation}
Adjusting the color values ensures that corresponding regions in the rendered images remain consistent within each clip. Furthermore, by refining overlapping frames between adjacent clips, we achieve smoother style transitions and improved temporal consistency throughout the video. Given that we adopt \modelname following a single-phase of video editing, \modelname (2nd stage) can be appropriately referred to as a single-phase refiner.
\begin{figure*}[!t]
\begin{center}
\includegraphics[width=0.85\linewidth]{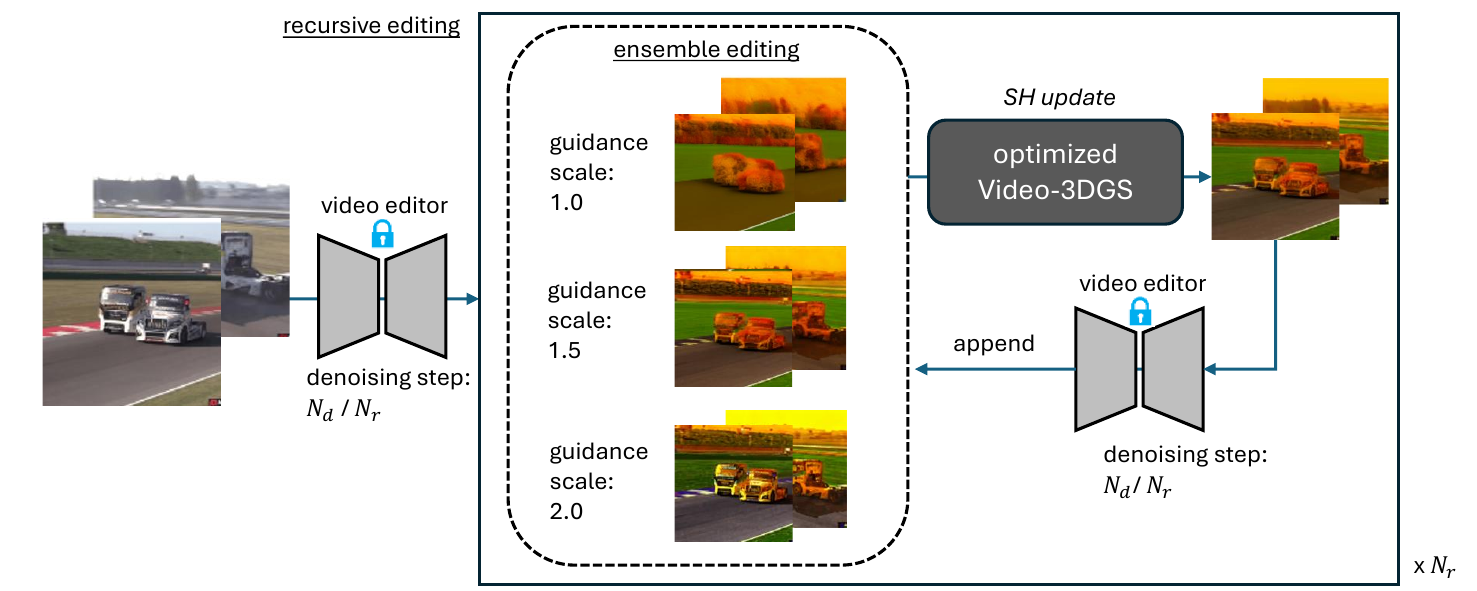}
\caption{
The overview of \modelname (2nd stage) as a plug-and-play refiner for video editing begins with fine-tuning the spherical coefficient of the optimized \modelname on an initially edited video, which is produced using an off-the-shelf video editor with the default hyperparameters: a denoising step $N_d$ and a guidance scale. This method, referred to as the single-phase refiner with 
\modelname, is further enhanced by our findings in ~\cref{fig:revisiting}. We split $N_d$ into a recursive number $N_r$
and fine-tune the spherical coefficient parameters against multiple outputs from varied guidance scales, aiming for improved temporal consistency and high fidelity to the editing text. This advanced approach is named the Recursive and Ensembled (RE) refinement. 
}
\label{fig:prgedit}
\end{center}
\end{figure*}

\textbf{Recursive and Ensembled Refiner}\quad 
In the single-phase refiner, we update $SH$ and $\sigma$ by minimizing the reconstruction loss between the rendered images and the edited frames. However, when the off-the-shelf video editor uses an excessive number of denoising steps ($N_d$) with editing prompts, some edited frames may lose essential structural details. To counteract this, we split $N_d$ into multiple phases (with $N_r=$ 2 for efficiency) and provide the detailed explanation for each phase.

\noindent\textit{First Phase:} we first apply the signle-phase refinement~\cref{eqn:single_edit} on the initially edited frames (e.g., after $N_d$/2 denoising steps). This produces a set of intermediate refined frames.

\noindent\textit{Initialization for the Next Phase:} we use the intermediate edited frames from above first phase as input for DDIM inversion (as in TokenFlow~\citep{tokenflow2023} and RAVE~\citep{2312.04524} or an image encoder (as in Text2Video-Zero~\citep{text2video-zero}) to initialize the subsequent phase.

\noindent\textit{Second Phase:} Then, we complete the remaining $N_d$/2 denoising steps for further video editing and follow with a final single-phase refinement on the newly edited frames.  

\noindent We refer to this overall procedure as \textbf{recursive refinement} with multiple phases. Additionally, the guidance parameters in the diffusion process (e.g., image guidance in Text2Video-Zero, text guidance in TokenFlow, and ControlNet guidance in RAVE) play a crucial role in the quality of the edits. As shown in~\cref{fig:revisiting}, different guidance scales can lead to significant variations. To reduce sensitivity to these scale changes and combine the benefits of multiple scales, we perform an ensemble update of $SH$ and $\sigma$ using edited videos generated at three different scales during each recursive phase. The ensemble of edits from one phase is then incorporated into the next, allowing each phase to benefit from previous refinements. For simplicity, we denote the 2nd-stage \modelname with \textbf{Recursive and Ensembled} refinement as \modelname (RE), as illustrated in~\cref{fig:prgedit}.

\section{Experimental Results}
\label{sec:results}
\vspace{-3mm}
In this section, we evaluate \modelname on two tasks: Video Reconstruction (\cref{sec:vr_sec}) and Video Editing (\cref{sec:ve_sec}). We provide the details of datasets and metrics used for those two tasks (\cref{sec:dataset_metrics}), ablation studies (\cref{sec:ablation}), and qualitative results (\cref{sec:more_vis}) in the Appendix.
\vspace{-1mm}


    

\begin{table}[t!]
  \centering
  \scalebox{0.7}{
  \begin{tabular}{c|c|c|c|c|c}
      \toprule
      \toprule
      \multicolumn{2}{c|}{\diagbox{Method}{Metrics}}& success rate & PSNR & SSIM & Time\\ \hline
      \multirow{4}{*}{NeRF-based} & Robust-Dyn ~\citep{liu2023robust} & 28 / 28 & 26.8 & 0.795 & 746m\\ 
       & NLA~\citep{kasten2021layered} & 28 / 28 & 27.3 & 0.795 & 309m\\
       & CoDeF~\citep{ouyang2023codef} & 28 / 28 & 29.4 & 0.869 & 28m \\ 
       & Nerv~\citep{chen2021nerv} & 28 / 28 & 34.6 & 0.980 & 20m \\ \hline
      \multirow{2}{*}{3DGS-based} & 3DGS~\citep{kerbl20233d} & 20 / 28 & 24.8 & 0.832 & 10m\\ 
      & Deform-3DGS~\citep{yang2023deformable3dgs} & 20 / 28 & 30.6 & 0.919 & 50m \\ \hline
       \multirow{3}{*}{3DGS-based (ours)} & Video-3DGS (3k) & 28 / 28 & 37.6 & 0.980  & 11m \\
       & Video-3DGS (5k) & 28 / 28 & 41.2 & 0.989  & 22m \\
       & Video-3DGS (10k) & 28 / 28 & 45.8 & 0.995 & 56m \\ 
      \bottomrule
      \bottomrule
    \end{tabular}
    }
    \vspace{1mm}
    \caption{Average  quantitative results of DAVIS videos for Video Reconstruction. `success rate' indicates the proportion of videos, out of the total 28 videos, that can be successfully reconstructed. Our method, \modelname, demonstrates efficient training and significantly outperforms NeRF-based and 3DGS-based methods.
    We report \modelname results trained with 3k, 5k, and 10k iterations.
     }
    \label{tab:recon_sum_quan}
\end{table}

\vspace{-1mm}





 \subsection{Video Reconstruction}
 \label{sec:vr_sec}
 \vspace{-2mm}
{\bf Baselines}\quad
As \modelname aims to reconstruct videos by learning scene representations, we  compare our approach with two representative methods: NeRF-based and 3DGS-based representations.
For NeRF-based methods, we select four state-of-the-art baselines: 1) NLA~\citep{kasten2021layered}, which  proposes a video reconstruction and editing method using layered atlases powered by the NeRF framework. 2) CoDEF~\citep{ouyang2023codef}, which leverages a canonical space of deformation fields to reconstruct and edit videos. 3) RobustDyn~\citep{liu2023robust}, an advanced view reconstruction and synthesis model that estimates camera poses in diverse settings. 4) Nerv~\cite{chen2021nerv}, which proposes neural representation to encode videos in neural networks.
For 3DGS-based methods, we first consider the original 3DGS method~\citep{kerbl20233d}, which lacks modules specific to video scenes.
Additionally, we select Deformable-3DGS~\citep{yang2023deformable3dgs}, a state-of-the-art approach that utilizes a deformation network on 3D Gaussian, serving as another strong baseline.





{\bf Quantitative Results}\quad
As illustrated in \cref{tab:recon_sum_quan}, our comprehensive video reconstruction experiments encompass 28 videos sourced from DAVIS.
We note that the four NeRF-based methods generally exhibit limitations in both reconstruction quality (measured by PSNR and SSIM) and efficiency (training time), primarily attributed to their implicit neural representation.
Conversely, the 3DGS method~\citep{kerbl20233d} demonstrates a significant reduction in training time  (less than 10 minutes in average of 20 videos), owing to its explicit 3D Gaussian representation and efficient rasterization.
However, the 3DGS method notably exhibits performance degradation (average PSNR of 20 videos: 24.8), as it is tailored for static scenes and lacks design considerations for dynamic video scenes.
On the other hand, the state-of-the-art baseline Deformable-3DGS, which employs a deformation field for time dimension on 3DGS, shows improved video reconstruction quality (average PSNR of 20 videos: 30.6); nevertheless, this enhancement comes at the expense of compromised training efficiency (50 minutes in average of 20 videos). Furthermore, due to the fundamental issue in COLMAP, they cannot conduct reconstruction on 8 videos, which hinders 3DGS from being used for wild video datasets. On the other hand, powered by the proposed MC-COLMAP and framework of Frg-3DGS/Bkg-3DGS, \modelname can reconstruct all 28 videos in high quality with shorter training time (\textit{iteration 3k}: \textbf{37.6 PSNR} with \textbf{11 minutes}; \textit{iteration 5k}: \textbf{41.2 PSNR} with \textbf{22 minutes}; \textit{iteration 10k}: \textbf{45.8 PSNR} with \textbf{56 minutes}; all results are measured by taking average over 28 videos).
Upon training for 3k iterations, \modelname significantly surpasses both the NeRF-based SoTA, Nerv~\citep{chen2021nerv}, and the 3DGS-based SoTA, Deform-3DGS~\citep{yang2023deformable3dgs}, in terms of video reconstruction quality and training efficiency. Specifically, it achieves an improvement in PSNR by +3 and +7 over Nerv and Deform-3DGS, respectively. Furthermore, \modelname demonstrates a notable improvement in training time efficiency, being 1.9 times faster than Nerv and 4.5 times faster than Deform-3DGS.

\begin{table*}
\centering
\resizebox{0.7\linewidth}{!}{
\begin{tabular}{l||cc|cc|cc}
\toprule
\toprule
 \multirow{2}{*}{\diagbox{Dataset}{Method}} & \multicolumn{2}{c|}{Text2Vid-Zero} & \multicolumn{2}{c|}{+\modelname} & \multicolumn{2}{c}{+\modelname (RE)}  \\
\cmidrule(lr){2-3} \cmidrule(lr){4-5} \cmidrule(lr){6-7} 
 & WarpSSIM↑ & $Q_{edit}$↑ &  WarpSSIM↑ & $Q_{edit}$↑ &  WarpSSIM↑ & $Q_{edit}$↑  \\

\midrule
 DAVIS & 0.691 & 20.1 & 0.827 (\textcolor{blue}{+13.6\%}) & 21.0 (\textcolor{blue}{+0.9\%}) & 0.899 (\textcolor{blue}{+20.8\%}) & 22.3 (\textcolor{blue}{+2.2\%})\\

Videovo &  0.773 & 21.9 & 0.902 (\textcolor{blue}{+12.9\%}) & 22.1 (\textcolor{blue}{+0.2\%}) & 0.926 (\textcolor{blue}{+15.3\%}) & 23.1 (\textcolor{blue}{+1.2\%})\\

Youtube &  0.701 & 19.8 & 0.885 (\textcolor{blue}{+18.4\%}) & 20.4 (\textcolor{blue}{+0.6\%}) & 0.922 (\textcolor{blue}{+22.1\%}) & 21.1 (\textcolor{blue}{+1.3\%}) \\

\bottomrule
\bottomrule

 \multirow{2}{*}{\diagbox{Dataset}{Method}} & \multicolumn{2}{c|}{TokenFlow} & \multicolumn{2}{c|}{+\modelname} & \multicolumn{2}{c}{+\modelname (RE)}  \\
\cmidrule(lr){2-3} \cmidrule(lr){4-5} \cmidrule(lr){6-7} 
 & WarpSSIM↑ & $Q_{edit}$↑ &  WarpSSIM↑ & $Q_{edit}$↑ &  WarpSSIM↑ & $Q_{edit}$↑  \\

\midrule
 DAVIS &   0.855 & 22.9 & 0.909 (\textcolor{blue}{+5.4\%}) & 23.9 (\textcolor{blue}{+1.0\%}) & 0.912 (\textcolor{blue}{+5.7\%}) & 24.8 (\textcolor{blue}{+1.9\%}) \\

Videovo &  0.897 & 22.6 & 0.933 (\textcolor{blue}{+4.6\%}) & 23.5 (\textcolor{blue}{+0.9\%}) & 0.937 (\textcolor{blue}{+4.9\%}) & 23.8 (\textcolor{blue}{+1.2\%})\\

Youtube &  0.848 & 22.1 & 0.923 (\textcolor{blue}{+7.5\%}) & 23.1 (\textcolor{blue}{+1.0\%}) & 0.923 (\textcolor{blue}{+7.5\%}) & 24.2 (\textcolor{blue}{+2.1\%})\\

\bottomrule
\bottomrule

 \multirow{2}{*}{\diagbox{Dataset}{Method}} & \multicolumn{2}{c|}{RAVE} & \multicolumn{2}{c|}{+\modelname} & \multicolumn{2}{c}{+\modelname (RE)}  \\
\cmidrule(lr){2-3} \cmidrule(lr){4-5} \cmidrule(lr){6-7} 
 & WarpSSIM↑ & $Q_{edit}$↑ &  WarpSSIM↑ & $Q_{edit}$↑ &  WarpSSIM↑ & $Q_{edit}$↑  \\

\midrule
 DAVIS & 0.872 & 23.4 & 0.908 (\textcolor{blue}{+3.6\%}) & 24.1 (\textcolor{blue}{+0.7\%}) & 0.913 (\textcolor{blue}{+4.1\%}) & 24.8 (\textcolor{blue}{+1.5\%})  \\

Videovo &  0.872 & 22.4 & 0.913 (\textcolor{blue}{+4.1\%}) & 23.5 (\textcolor{blue}{+1.1\%}) & 0.923 (\textcolor{blue}{+5.1\%}) & 23.6 (\textcolor{blue}{+1.2\%})  \\

Youtube & 0.855 & 21.7 & 0.918 (\textcolor{blue}{+6.3\%}) & 22.9 (\textcolor{blue}{+1.2\%}) & 0.921 (\textcolor{blue}{+6.6\%}) & 23.5  (\textcolor{blue}{+1.8\%})  \\

\bottomrule
\bottomrule

\end{tabular}
}
\caption{Quantitative results  of Video Editing on three different datasets: DAVIS, Videvo, Youtube. Each dataset contains four editing categories: style change, object change, background change, multiple change.
We present the results for average WarpSSIM and $Q_{edit}$ of four categories,
with text color indicating the effect of \modelname.
Our \modelname increases both temporal consistency and overall video editing performance across all initial video editors. ↑: the higher, the better.
}
\label{tab:edit_quan}
\end{table*}



\begin{figure}[t!]
  \centering
  \scalebox{0.55}{
  \begin{tikzpicture}
    \begin{axis}[
        ybar,
        symbolic x coords={Text2Video-Zero, TokenFlow, RAVE},
        xtick=data,
        nodes near coords,
        nodes near coords align={vertical},
        ymin=0,ymax=1,
        ylabel={User Preference},
        bar width=15pt,
        enlarge x limits=0.19,
        legend style={at={(0.5,-0.15)},
        anchor=north,legend columns=-1},
        /tikz/every even column/.append style={column sep=0.5cm}
    ]
    \addplot[fill=orange] coordinates {(Text2Video-Zero,0.18) (TokenFlow,0.23) (RAVE,0.2)};
    \addplot[fill=cyan!98] coordinates {(Text2Video-Zero,0.72) (TokenFlow,0.40) (RAVE,0.55)};
    \addplot[fill=lightgray] coordinates {(Text2Video-Zero,0.1) (TokenFlow,0.37) (RAVE,0.25)};
    \legend{Baseline , w/ Video-3DGS, Equal preference}
    \end{axis}
  \end{tikzpicture}
  }
  \scalebox{0.55}{
  \begin{tikzpicture}
    \begin{axis}[
        ybar,
        symbolic x coords={Text2Video-Zero, TokenFlow, RAVE},
        xtick=data,
        nodes near coords,
        nodes near coords align={vertical},
        ymin=0,ymax=1,
        ylabel={User Preference},
        bar width=15pt,
        enlarge x limits=0.19,
        legend style={at={(0.5,-0.15)},
        anchor=north,legend columns=-1},
        /tikz/every even column/.append style={column sep=0.5cm}
    ]
    \addplot[fill=cyan!98] coordinates {(Text2Video-Zero,0.21) (TokenFlow,0.27) (RAVE,0.25)};
    \addplot[fill=blue] coordinates {(Text2Video-Zero,0.69) (TokenFlow,0.35) (RAVE,0.53)};
    \addplot[fill=lightgray] coordinates {(Text2Video-Zero,0.1) (TokenFlow,0.38) (RAVE,0.22)};
    \legend{w/ Video-3DGS, w/ Video-3DGS (RE), Equal preference}
    \end{axis}
  \end{tikzpicture}
  }
  \caption{User study of Video-3DGS on three video editors. 
  }
  \label{fig:user_study}
  \vspace{-4mm}
\end{figure}
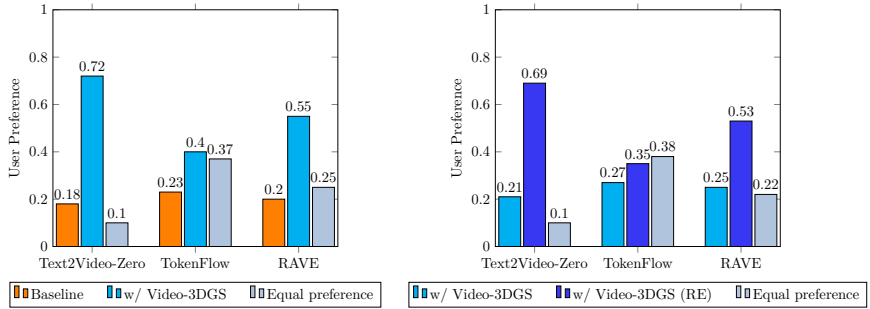

\vspace{-2mm}

 \subsection{Video Editing}
\label{sec:ve_sec}
\vspace{-2mm}

{\bf Importance of 3D Gaussians for Video Editing Task}\quad
The primary goal of \modelname is to enhance the temporal consistency of video editing by leveraging 3D Gaussian representations with a two-stage approach. Specifically, in the 1st stage, our \modelname reconstructs videos using 3D Gaussians to establish temporal correspondence between frames within each clip and link neighboring clips with overlapping frames. It motivates us to prove the video reconstruction capability of \modelname, showcasing its ability to preserve the content and structure of original raw video without need of any 3D task exploration. Afterwards, we are able to exploit this reconstruction ability for video editing tasks in the 2nd stage. We provide explicit evidence of 3D Gaussians’ suitability for temporal correspondence as shown in this video \href{https://anonymous-video-3dgs-time-correspond.github.io}{link}, where we visualize the correspondence for the first five groups of 3D Gaussians. As shown in the visualization, \modelname ensures the correspondences of similar areas across frames using 3D Gaussians of the same group. We can enhance this further by incorporating the visualization of pixel trajectories tracked by the 3D Gaussians. This enhancement is demonstrated in the provided video \href{https://anonymous-video-3dgs-tracking.github.io/}{link}.

{\bf Baselines}\quad
We meticulously choose three zero-shot video editing methods as baseline comparisons.
1) Text2Video-Zero~\citep{text2video-zero}, which extends Instruct-pix2pix~\citep{brooks2022instructpix2pix} to the video domain by inserting temporal attention within the diffusion model.
2) TokenFlow~\citep{tokenflow2023}, which achieves temporal smoothness through the propagation of diffusion features using inter-frame correspondences.
3) RAVE~\citep{2312.04524}, which employs a noise-shuffling strategy and grid trick for enhancing video editing capabilities.



{\bf Quantitative Results}\quad
We assessed \modelname's editing capability on top of those three zero-shot video editors in~\cref{tab:edit_quan}.
\modelname significantly enhances temporal consistency (WarpSSIM score) across various editing scenarios spanning three different datasets. Consequently, this improvement generally yields superior final video editing results ($Q_{edit}$). It proves that \modelname generally provides robust temporal editing guidance to existing video editors. We also observe that our proposed recursive and ensembled refinement can further improve both temporal consistency and video editing results on all of video editors and datasets.








{\bf User Study}\quad
To explore user preferences for video editing outcomes, we conducted a user study detailed in~\cref{fig:user_study}.
We enlisted 20 participants and presented them with 30 randomly paired videos (10 pairs for each of the adopted off-the-shelf video editors) derived from the LOVEU video results.
Participants were tasked with selecting their preferred edited video, taking into account fidelity to the provided text prompt (for editing) and temporal consistency. Consistent with the findings in the left panel of~\cref{fig:user_study}, we observed that users generally favored \modelname guided video editing results over the original edited videos. Moreover, in the right panel of~\cref{fig:user_study}, we conducted another user study to show the effectiveness of recursive and ensembled refinement in editing with \modelname.

\vspace{-2mm}



\section{Discussion \& Limitations}
\label{sec:limitations}
While the proposed \modelname exhibits remarkable performance in both video reconstruction and editing tasks, certain challenges persist. 

\noindent In video reconstruction, our \modelname still relies on the success of MC-COLMAP even though it is more robust version compared to original COLMAP. Following are scenarios when MC-COLMAP can struggle or fail. \textit{Significant Occlusions of Foreground Objects}: When a foreground object becomes fully or partially occluded in consecutive frames, MC-COLMAP’s decomposition process can struggle to track and re-identify the object. \textit{Excessive Motion Blur}: Rapid camera or object movements can induce motion blur, making it difficult for feature detection algorithms to match points across frames accurately. If the blurred regions prevent consistent feature extraction, the reconstruction pipeline may fail to build a coherent model for certain frames or produce large alignment errors. We further acknowledge that MC-COLMAP for \modelname requires a preprocessing step for foreground object extraction using an off-the-shelf panoptic segmenter. However, this is notably simpler compared to the five-stage preprocessing pipeline of shape-of-motion~\cite{som2024}, which involves mask estimation, metric depth, monocular depth, camera estimation, and 2D tracking. Furthermore, we studied the influence of different variants and settings of the off-the-shelf segmenter on video reconstruction quality to ensure robustness as we discuss more in ``Ablations on Segmentation Modules'' of \cref{sec:ablation_first}.

In video editing, unsatisfactory results may arise if the underlying video editor fails or if significant changes to object shapes are required, which contradicts the inherent property of 3D Gaussians prioritizing the preservation of original structure.
As a future direction for \modelname, we aim to develop an architecture capable of manipulating 3D Gaussians in a more flexible manner to accommodate desired shape changes.
Nevertheless, our work represents a pioneering effort in extending 3DGS to enhance the temporal consistency of initially edited videos from any zero-shot video editor. We envision that this will inspire further research in a zero-shot video editing community.

Unlike other conventional neural representation papers, our work does not explore the task of novel view synthesis. Instead, our method focuses on video reconstruction and editing by encoding and fitting videos with 3DGS. We anticipate that our \modelname will also serve as a fundamental framework for 4D novel view synthesis in future research. 

\section{Conclusion}
\label{sec:conclusion}


We introduced \textbf{\modelname}, a novel two-stage approach tailored to edit dynamic monocular video scenes by reconstructing them with 3D Gaussian Splatting (3DGS).
The first stage of \modelname lies its capacity to spatially and temporally decompose videos, streamlining motion representation across clips.
It surpasses previous state-of-the-art methods in video reconstruction. The second stage of \modelname leverages the reconstruction ability to refine the edited videos from three different zero-shot video editors, which enhances both temporal consistency and editing results.
We anticipate significant positive societal impacts from our method, as it consistently performs well in representing diverse videos and refining outputs from zero-shot video editors. We expect it to benefit various applications, such as Entertainment with video synthesis and AR/VR.

\noindent \textbf{Acknowledgement}
This work was supported in part by the Institute of Information and Communications Technology Planning and Evaluation (IITP) grant funded by the Korea Government (MSIT) (Artificial Intelligence Innovation Hub) under Grant 2021-0-02068. This research was also partially supported by the KAIST Cross-Generation Collaborative Lab Project.

\bibliography{main}

\begin{thebibliography}{62}
\providecommand{\natexlab}[1]{#1}
\providecommand{\url}[1]{\texttt{#1}}
\expandafter\ifx\csname urlstyle\endcsname\relax
  \providecommand{\doi}[1]{doi: #1}\else
  \providecommand{\doi}{doi: \begingroup \urlstyle{rm}\Url}\fi

\bibitem[Andonian et~al.(2021)Andonian, Osmany, Cui, Park, Jahanian, Torralba, and Bau]{bau2021paintbyword}
Alex Andonian, Sabrina Osmany, Audrey Cui, YeonHwan Park, Ali Jahanian, Antonio Torralba, and David Bau.
\newblock Paint by word.
\newblock \emph{arXiv preprint arXiv:2103.10951}, 2021.

\bibitem[Avrahami et~al.(2022)Avrahami, Lischinski, and Fried]{Avrahami_2022_CVPR}
Omri Avrahami, Dani Lischinski, and Ohad Fried.
\newblock Blended diffusion for text-driven editing of natural images.
\newblock In \emph{CVPR}, 2022.

\bibitem[Bai et~al.(2024)Bai, He, Wang, Guo, Hu, Liu, and Bian]{bai2024uniedit}
Jianhong Bai, Tianyu He, Yuchi Wang, Junliang Guo, Haoji Hu, Zuozhu Liu, and Jiang Bian.
\newblock Uniedit: A unified tuning-free framework for video motion and appearance editing.
\newblock \emph{arXiv preprint arXiv:2402.13185}, 2024.

\bibitem[Barron et~al.(2021)Barron, Mildenhall, Tancik, Hedman, Martin-Brualla, and Srinivasan]{barron2021mipnerf}
Jonathan~T Barron, Ben Mildenhall, Matthew Tancik, Peter Hedman, Ricardo Martin-Brualla, and Pratul~P Srinivasan.
\newblock Mip-nerf: A multiscale representation for anti-aliasing neural radiance fields.
\newblock In \emph{CVPR}, 2021.

\bibitem[Brooks et~al.(2023)Brooks, Holynski, and Efros]{brooks2022instructpix2pix}
Tim Brooks, Aleksander Holynski, and Alexei~A Efros.
\newblock Instructpix2pix: Learning to follow image editing instructions.
\newblock In \emph{CVPR}, 2023.

\bibitem[Cao \& Johnson(2023)Cao and Johnson]{cao2023hexplane}
Ang Cao and Justin Johnson.
\newblock Hexplane: A fast representation for dynamic scenes.
\newblock In \emph{CVPR}, 2023.

\bibitem[Ceylan et~al.(2023)Ceylan, Huang, and Mitra]{ceylan2023pix2video}
Duygu Ceylan, Chun-Hao Huang, and Niloy~J. Mitra.
\newblock Pix2video: Video editing using image diffusion.
\newblock In \emph{ICCV}, 2023.

\bibitem[Chen et~al.(2021{\natexlab{a}})Chen, Xu, Zhao, Zhang, Xiang, Yu, and Su]{chen2021mvsnerf}
Anpei Chen, Zexiang Xu, Fuqiang Zhao, Xiaoshuai Zhang, Fanbo Xiang, Jingyi Yu, and Hao Su.
\newblock Mvsnerf: Fast generalizable radiance field reconstruction from multi-view stereo.
\newblock \emph{arXiv preprint arXiv:2103.15595}, 2021{\natexlab{a}}.

\bibitem[Chen et~al.(2021{\natexlab{b}})Chen, He, Wang, Ren, Lim, and Shrivastava]{chen2021nerv}
Hao Chen, Bo~He, Hanyu Wang, Yixuan Ren, Ser~Nam Lim, and Abhinav Shrivastava.
\newblock Nerv: Neural representations for videos.
\newblock \emph{NeurIPS}, 2021{\natexlab{b}}.

\bibitem[Cheng et~al.(2023)Cheng, Oh, Price, Schwing, and Lee]{cheng2023tracking}
Ho~Kei Cheng, Seoung~Wug Oh, Brian Price, Alexander Schwing, and Joon-Young Lee.
\newblock Tracking anything with decoupled video segmentation.
\newblock In \emph{ICCV}, 2023.

\bibitem[Chng et~al.(2022)Chng, Ramasinghe, Sherrah, and Lucey]{chng2022garf}
Shin-Fang Chng, Sameera Ramasinghe, Jamie Sherrah, and Simon Lucey.
\newblock Garf: Gaussian activated radiance fields for high fidelity reconstruction and pose estimation.
\newblock \emph{arXiv preprint arXiv:2204.05735}, 2022.

\bibitem[Couairon et~al.(2022)Couairon, Verbeek, Schwenk, and Cord]{couairon2022diffedit}
Guillaume Couairon, Jakob Verbeek, Holger Schwenk, and Matthieu Cord.
\newblock Diffedit: Diffusion-based semantic image editing with mask guidance.
\newblock \emph{arXiv preprint arXiv:2210.11427}, 2022.

\bibitem[Fan et~al.(2023)Fan, Wang, Wen, Zhu, Xu, and Wang]{fan2023lightgaussian}
Zhiwen Fan, Kevin Wang, Kairun Wen, Zehao Zhu, Dejia Xu, and Zhangyang Wang.
\newblock Lightgaussian: Unbounded 3d gaussian compression with 15x reduction and 200+ fps.
\newblock \emph{arXiv preprint arXiv:2311.17245}, 2023.

\bibitem[Gal et~al.(2022)Gal, Alaluf, Atzmon, Patashnik, Bermano, Chechik, and Cohen-Or]{gal2022textual}
Rinon Gal, Yuval Alaluf, Yuval Atzmon, Or~Patashnik, Amit~H. Bermano, Gal Chechik, and Daniel Cohen-Or.
\newblock An image is worth one word: Personalizing text-to-image generation using textual inversion.
\newblock \emph{arXiv preprint arXiv:2208.01618}, 2022.

\bibitem[Garbin et~al.(2021)Garbin, Kowalski, Johnson, Shotton, and Valentin]{garbin2021fastnerf}
Stephan~J. Garbin, Marek Kowalski, Matthew Johnson, Jamie Shotton, and Julien Valentin.
\newblock Fastnerf: High-fidelity neural rendering at 200fps.
\newblock \emph{arXiv preprint arXiv:2103.10380}, 2021.

\bibitem[Geyer et~al.(2023)Geyer, Bar-Tal, Bagon, and Dekel]{tokenflow2023}
Michal Geyer, Omer Bar-Tal, Shai Bagon, and Tali Dekel.
\newblock Tokenflow: Consistent diffusion features for consistent video editing.
\newblock \emph{arXiv preprint arXiv:2307.10373}, 2023.

\bibitem[Gu et~al.(2021)Gu, Liu, Wang, and Theobalt]{gu2021stylenerf}
Jiatao Gu, Lingjie Liu, Peng Wang, and Christian Theobalt.
\newblock Stylenerf: A style-based 3d-aware generator for high-resolution image synthesis.
\newblock \emph{arXiv preprint arXiv:2110.08985}, 2021.

\bibitem[Hertz et~al.(2022)Hertz, Mokady, Tenenbaum, Aberman, Pritch, and Cohen-Or]{hertz2022prompttoprompt}
Amir Hertz, Ron Mokady, Jay Tenenbaum, Kfir Aberman, Yael Pritch, and Daniel Cohen-Or.
\newblock Prompt-to-prompt image editing with cross attention control.
\newblock \emph{arXiv preprint arXiv:2208.01626}, 2022.

\bibitem[Ho et~al.(2020)Ho, Jain, and Abbeel]{ho2020denoising}
Jonathan Ho, Ajay Jain, and Pieter Abbeel.
\newblock Denoising diffusion probabilistic models.
\newblock \emph{NeurIPS}, 2020.

\bibitem[Ho et~al.(2022)Ho, Salimans, Gritsenko, Chan, Norouzi, and Fleet]{ho2022video}
Jonathan Ho, Tim Salimans, Alexey Gritsenko, William Chan, Mohammad Norouzi, and David~J Fleet.
\newblock Video diffusion models.
\newblock \emph{NeurIPS}, 2022.

\bibitem[Jeong et~al.(2021)Jeong, Ahn, Choy, Anandkumar, Cho, and Park]{jeong2021selfcalibrating}
Yoonwoo Jeong, Seokjun Ahn, Christopher Choy, Animashree Anandkumar, Minsu Cho, and Jaesik Park.
\newblock Self-calibrating neural radiance fields.
\newblock \emph{arXiv preprint arXiv:2108.13826}, 2021.

\bibitem[Kara et~al.(2023)Kara, Kurtkaya, Yesiltepe, Rehg, and Yanardag]{2312.04524}
Ozgur Kara, Bariscan Kurtkaya, Hidir Yesiltepe, James~M. Rehg, and Pinar Yanardag.
\newblock Rave: Randomized noise shuffling for fast and consistent video editing with diffusion models.
\newblock \emph{arXiv preprint arXiv:2312.04524}, 2023.

\bibitem[Karnewar et~al.(2022)Karnewar, Ritschel, Wang, and Mitra]{Karnewar_2022}
Animesh Karnewar, Tobias Ritschel, Oliver Wang, and Niloy Mitra.
\newblock Relu fields: The little non-linearity that could.
\newblock In \emph{SIGGRAPH}, 2022.

\bibitem[Kasten et~al.(2021)Kasten, Ofri, Wang, and Dekel]{kasten2021layered}
Yoni Kasten, Dolev Ofri, Oliver Wang, and Tali Dekel.
\newblock Layered neural atlases for consistent video editing.
\newblock \emph{ACM Transactions on Graphics (TOG)}, 40\penalty0 (6):\penalty0 1--12, 2021.

\bibitem[Kerbl et~al.(2023)Kerbl, Kopanas, Leimk{\"u}hler, and Drettakis]{kerbl20233d}
Bernhard Kerbl, Georgios Kopanas, Thomas Leimk{\"u}hler, and George Drettakis.
\newblock 3d gaussian splatting for real-time radiance field rendering.
\newblock \emph{ACM Transactions on Graphics}, 42\penalty0 (4), 2023.

\bibitem[Khachatryan et~al.(2023)Khachatryan, Movsisyan, Tadevosyan, Henschel, Wang, Navasardyan, and Shi]{text2video-zero}
Levon Khachatryan, Andranik Movsisyan, Vahram Tadevosyan, Roberto Henschel, Zhangyang Wang, Shant Navasardyan, and Humphrey Shi.
\newblock Text2video-zero: Text-to-image diffusion models are zero-shot video generators.
\newblock \emph{arXiv preprint arXiv:2303.13439}, 2023.

\bibitem[Ku et~al.(2024)Ku, Wei, Ren, Yang, and Chen]{ku2024anyv2v}
Max Ku, Cong Wei, Weiming Ren, Huan Yang, and Wenhu Chen.
\newblock Anyv2v: A plug-and-play framework for any video-to-video editing tasks.
\newblock \emph{arXiv preprint arXiv:2403.14468}, 2024.

\bibitem[Kundu et~al.(2022)Kundu, Genova, Yin, Fathi, Pantofaru, Guibas, Tagliasacchi, Dellaert, and Funkhouser]{kundu2022panoptic}
Abhijit Kundu, Kyle Genova, Xiaoqi Yin, Alireza Fathi, Caroline Pantofaru, Leonidas Guibas, Andrea Tagliasacchi, Frank Dellaert, and Thomas Funkhouser.
\newblock Panoptic neural fields: A semantic object-aware neural scene representation.
\newblock \emph{arXiv preprint arXiv:2205.04334}, 2022.

\bibitem[Lee et~al.(2024)Lee, Lee, Sun, Ali, and Park]{lee2024deblurring}
Byeonghyeon Lee, Howoong Lee, Xiangyu Sun, Usman Ali, and Eunbyung Park.
\newblock Deblurring 3d gaussian splatting.
\newblock \emph{arXiv preprint arXiv:2401.00834}, 2024.

\bibitem[Li et~al.(2023)Li, Li, Savarese, and Hoi]{li2023blip}
Junnan Li, Dongxu Li, Silvio Savarese, and Steven Hoi.
\newblock Blip-2: Bootstrapping language-image pre-training with frozen image encoders and large language models.
\newblock \emph{arXiv preprint arXiv:2301.12597}, 2023.

\bibitem[Lin et~al.(2021)Lin, Ma, Torralba, and Lucey]{lin2021barf}
Chen-Hsuan Lin, Wei-Chiu Ma, Antonio Torralba, and Simon Lucey.
\newblock Barf: Bundle-adjusting neural radiance fields.
\newblock \emph{arXiv preprint arXiv:2104.06405}, 2021.

\bibitem[Liu et~al.(2023)Liu, Gao, Meuleman, Tseng, Saraf, Kim, Chuang, Kopf, and Huang]{liu2023robust}
Yu-Lun Liu, Chen Gao, Andreas Meuleman, Hung-Yu Tseng, Ayush Saraf, Changil Kim, Yung-Yu Chuang, Johannes Kopf, and Jia-Bin Huang.
\newblock Robust dynamic radiance fields.
\newblock In \emph{CVPR}, 2023.

\bibitem[Ma et~al.(2022)Ma, Li, Liao, Zhang, Wang, Wang, and Sander]{ma2022deblurnerf}
Li~Ma, Xiaoyu Li, Jing Liao, Qi~Zhang, Xuan Wang, Jue Wang, and Pedro~V. Sander.
\newblock Deblur-nerf: Neural radiance fields from blurry images.
\newblock \emph{arXiv preprint arXiv:2111.14292}, 2022.

\bibitem[Martin-Brualla et~al.(2021)Martin-Brualla, Radwan, Sajjadi, Barron, Dosovitskiy, and Duckworth]{martinbrualla2021nerf}
Ricardo Martin-Brualla, Noha Radwan, Mehdi S.~M. Sajjadi, Jonathan~T. Barron, Alexey Dosovitskiy, and Daniel Duckworth.
\newblock Nerf in the wild: Neural radiance fields for unconstrained photo collections.
\newblock \emph{arXiv preprint arXiv:2008.02268}, 2021.

\bibitem[Meng et~al.(2022)Meng, He, Song, Song, Wu, Zhu, and Ermon]{meng2022sdedit}
Chenlin Meng, Yutong He, Yang Song, Jiaming Song, Jiajun Wu, Jun-Yan Zhu, and Stefano Ermon.
\newblock Sdedit: Guided image synthesis and editing with stochastic differential equations.
\newblock \emph{arXiv preprint arXiv:2108.01073}, 2022.

\bibitem[Meuleman et~al.(2023)Meuleman, Liu, Gao, Huang, Kim, Kim, and Kopf]{meuleman2023localrf}
Andreas Meuleman, Yu-Lun Liu, Chen Gao, Jia-Bin Huang, Changil Kim, Min~H. Kim, and Johannes Kopf.
\newblock Progressively optimized local radiance fields for robust view synthesis.
\newblock In \emph{CVPR}, 2023.

\bibitem[Mildenhall et~al.(2021)Mildenhall, Srinivasan, Tancik, Barron, Ramamoorthi, and Ng]{mildenhall2021nerf}
Ben Mildenhall, Pratul~P Srinivasan, Matthew Tancik, Jonathan~T Barron, Ravi Ramamoorthi, and Ren Ng.
\newblock Nerf: Representing scenes as neural radiance fields for view synthesis.
\newblock \emph{Communications of the ACM}, 65\penalty0 (1):\penalty0 99--106, 2021.

\bibitem[M\"uller(2021)]{tiny_cuda_nn}
Thomas M\"uller.
\newblock {tiny-cuda-nn}, 4 2021.
\newblock URL \url{https://github.com/NVlabs/tiny-cuda-nn}.

\bibitem[Ouyang et~al.(2023)Ouyang, Wang, Xiao, Bai, Zhang, Zheng, Zhou, Chen, and Shen]{ouyang2023codef}
Hao Ouyang, Qiuyu Wang, Yuxi Xiao, Qingyan Bai, Juntao Zhang, Kecheng Zheng, Xiaowei Zhou, Qifeng Chen, and Yujun Shen.
\newblock Codef: Content deformation fields for temporally consistent video processing.
\newblock \emph{arXiv preprint arXiv:2308.07926}, 2023.

\bibitem[Paszke et~al.(2019)Paszke, Gross, Massa, Lerer, Bradbury, Chanan, Killeen, Lin, Gimelshein, Antiga, et~al.]{paszke2019pytorch}
Adam Paszke, Sam Gross, Francisco Massa, Adam Lerer, James Bradbury, Gregory Chanan, Trevor Killeen, Zeming Lin, Natalia Gimelshein, Luca Antiga, et~al.
\newblock Pytorch: An imperative style, high-performance deep learning library.
\newblock \emph{NeurIPS}, 2019.

\bibitem[Pont-Tuset et~al.(2017)Pont-Tuset, Perazzi, Caelles, Arbel\'aez, Sorkine-Hornung, and {Van Gool}]{Pont-Tuset_arXiv_2017}
Jordi Pont-Tuset, Federico Perazzi, Sergi Caelles, Pablo Arbel\'aez, Alexander Sorkine-Hornung, and Luc {Van Gool}.
\newblock The 2017 davis challenge on video object segmentation.
\newblock \emph{arXiv preprint arXiv:1704.00675}, 2017.

\bibitem[Qi et~al.(2023)Qi, Cun, Zhang, Lei, Wang, Shan, and Chen]{qi2023fatezero}
Chenyang Qi, Xiaodong Cun, Yong Zhang, Chenyang Lei, Xintao Wang, Ying Shan, and Qifeng Chen.
\newblock Fatezero: Fusing attentions for zero-shot text-based video editing.
\newblock \emph{arXiv preprint arXiv:2303.09535}, 2023.

\bibitem[Reiser et~al.(2021)Reiser, Peng, Liao, and Geiger]{reiser2021kilonerf}
Christian Reiser, Songyou Peng, Yiyi Liao, and Andreas Geiger.
\newblock Kilonerf: Speeding up neural radiance fields with thousands of tiny mlps.
\newblock \emph{arXiv preprint arXiv:2103.13744}, 2021.

\bibitem[Rombach et~al.(2022)Rombach, Blattmann, Lorenz, Esser, and Ommer]{rombach2022high}
Robin Rombach, Andreas Blattmann, Dominik Lorenz, Patrick Esser, and Bj{\"o}rn Ommer.
\newblock High-resolution image synthesis with latent diffusion models.
\newblock In \emph{CVPR}, 2022.

\bibitem[Ruiz et~al.(2022)Ruiz, Li, Jampani, Pritch, Rubinstein, and Aberman]{ruiz2022dreambooth}
Nataniel Ruiz, Yuanzhen Li, Varun Jampani, Yael Pritch, Michael Rubinstein, and Kfir Aberman.
\newblock Dreambooth: Fine tuning text-to-image diffusion models for subject-driven generation.
\newblock \emph{arXiv preprint arxiv:2208.12242}, 2022.

\bibitem[Sch\"{o}nberger \& Frahm(2016)Sch\"{o}nberger and Frahm]{schoenberger2016sfm}
Johannes~Lutz Sch\"{o}nberger and Jan-Michael Frahm.
\newblock Structure-from-motion revisited.
\newblock In \emph{CVPR}, 2016.

\bibitem[Singer et~al.(2022)Singer, Polyak, Hayes, Yin, An, Zhang, Hu, Yang, Ashual, Gafni, et~al.]{singer2022make}
Uriel Singer, Adam Polyak, Thomas Hayes, Xi~Yin, Jie An, Songyang Zhang, Qiyuan Hu, Harry Yang, Oron Ashual, Oran Gafni, et~al.
\newblock Make-a-video: Text-to-video generation without text-video data.
\newblock \emph{arXiv preprint arXiv:2209.14792}, 2022.

\bibitem[Sohl-Dickstein et~al.(2015)Sohl-Dickstein, Weiss, Maheswaranathan, and Ganguli]{sohl2015deep}
Jascha Sohl-Dickstein, Eric Weiss, Niru Maheswaranathan, and Surya Ganguli.
\newblock Deep unsupervised learning using nonequilibrium thermodynamics.
\newblock In \emph{ICML}, 2015.

\bibitem[Song et~al.(2020)Song, Meng, and Ermon]{song2020denoising}
Jiaming Song, Chenlin Meng, and Stefano Ermon.
\newblock Denoising diffusion implicit models.
\newblock \emph{arXiv preprint arXiv:2010.02502}, 2020.

\bibitem[Teed \& Deng(2020)Teed and Deng]{teed2020raft}
Zachary Teed and Jia Deng.
\newblock Raft: Recurrent all-pairs field transforms for optical flow.
\newblock In \emph{ECCV}, 2020.

\bibitem[Wang et~al.(2022)Wang, Cui, Salcudean, and Wang]{wang2022generalizable}
Dan Wang, Xinrui Cui, Septimiu Salcudean, and Z.~Jane Wang.
\newblock Generalizable neural radiance fields for novel view synthesis with transformer.
\newblock \emph{arXiv preprint arXiv:2206.05375}, 2022.

\bibitem[Wang et~al.(2024{\natexlab{a}})Wang, Ye, Gao, Zeng, Austin, Li, and Kanazawa]{som2024}
Qianqian Wang, Vickie Ye, Hang Gao, Weijia Zeng, Jake Austin, Zhengqi Li, and Angjoo Kanazawa.
\newblock Shape of motion: 4d reconstruction from a single video.
\newblock \emph{arXiv preprint arXiv:2407.13764}, 2024{\natexlab{a}}.

\bibitem[Wang et~al.(2024{\natexlab{b}})Wang, Chen, Ma, Zhou, Huang, Wang, Yang, He, Yu, Yang, et~al.]{wang2024lavie}
Yaohui Wang, Xinyuan Chen, Xin Ma, Shangchen Zhou, Ziqi Huang, Yi~Wang, Ceyuan Yang, Yinan He, Jiashuo Yu, Peiqing Yang, et~al.
\newblock Lavie: High-quality video generation with cascaded latent diffusion models.
\newblock \emph{International Journal of Computer Vision}, pp.\  1--20, 2024{\natexlab{b}}.

\bibitem[Wiles et~al.(2020)Wiles, Gkioxari, Szeliski, and Johnson]{wiles2020synsin}
Olivia Wiles, Georgia Gkioxari, Richard Szeliski, and Justin Johnson.
\newblock Synsin: End-to-end view synthesis from a single image.
\newblock \emph{arXiv preprint arXiv:1912.08804}, 2020.

\bibitem[Wu et~al.(2023{\natexlab{a}})Wu, Yi, Fang, Xie, Zhang, Wei, Liu, Tian, and Wang]{wu20234d}
Guanjun Wu, Taoran Yi, Jiemin Fang, Lingxi Xie, Xiaopeng Zhang, Wei Wei, Wenyu Liu, Qi~Tian, and Xinggang Wang.
\newblock 4d gaussian splatting for real-time dynamic scene rendering.
\newblock \emph{arXiv preprint arXiv:2310.08528}, 2023{\natexlab{a}}.

\bibitem[Wu et~al.(2023{\natexlab{b}})Wu, Li, Gao, Dong, Bai, Singh, Xiang, Li, Huang, Sun, He, Hu, Hu, Huang, Zhu, Cheng, Tang, Shou, Keutzer, and Iandola]{wu2023cvpr}
Jay~Zhangjie Wu, Xiuyu Li, Difei Gao, Zhen Dong, Jinbin Bai, Aishani Singh, Xiaoyu Xiang, Youzeng Li, Zuwei Huang, Yuanxi Sun, Rui He, Feng Hu, Junhua Hu, Hai Huang, Hanyu Zhu, Xu~Cheng, Jie Tang, Mike~Zheng Shou, Kurt Keutzer, and Forrest Iandola.
\newblock Cvpr 2023 text guided video editing competition.
\newblock \emph{arXiv preprint arXiv:2310.16003}, 2023{\natexlab{b}}.

\bibitem[Yang et~al.(2023)Yang, Gao, Zhou, Jiao, Zhang, and Jin]{yang2023deformable3dgs}
Ziyi Yang, Xinyu Gao, Wen Zhou, Shaohui Jiao, Yuqing Zhang, and Xiaogang Jin.
\newblock Deformable 3d gaussians for high-fidelity monocular dynamic scene reconstruction.
\newblock \emph{arXiv preprint arXiv:2309.13101}, 2023.

\bibitem[Yu et~al.(2021)Yu, Li, Tancik, Li, Ng, and Kanazawa]{yu2021plenoctrees}
Alex Yu, Ruilong Li, Matthew Tancik, Hao Li, Ren Ng, and Angjoo Kanazawa.
\newblock Plenoctrees for real-time rendering of neural radiance fields.
\newblock \emph{arXiv preprint arXiv:2103.14024}, 2021.

\bibitem[Yu et~al.(2023)Yu, Chen, Huang, Sattler, and Geiger]{Yu2023MipSplatting}
Zehao Yu, Anpei Chen, Binbin Huang, Torsten Sattler, and Andreas Geiger.
\newblock Mip-splatting: Alias-free 3d gaussian splatting.
\newblock \emph{arXiv preprint arXiv:2311.16493}, 2023.

\bibitem[Zhang et~al.(2023)Zhang, Rao, and Agrawala]{zhang2023adding}
Lvmin Zhang, Anyi Rao, and Maneesh Agrawala.
\newblock Adding conditional control to text-to-image diffusion models.
\newblock In \emph{ICCV}, 2023.

\bibitem[Zhi et~al.(2021)Zhi, Laidlow, Leutenegger, and Davison]{Zhi:etal:ICCV2021}
Shuaifeng Zhi, Tristan Laidlow, Stefan Leutenegger, and Andrew~J. Davison.
\newblock In-place scene labelling and understanding with implicit scene representation.
\newblock In \emph{ICCV}, 2021.

\bibitem[Zhou et~al.(2023)Zhou, Kim, Wang, Florence, and Finn]{zhou2023nerf}
Allan Zhou, Moo~Jin Kim, Lirui Wang, Pete Florence, and Chelsea Finn.
\newblock Nerf in the palm of your hand: Corrective augmentation for robotics via novel-view synthesis.
\newblock \emph{arXiv preprint arXiv:2301.08556}, 2023.

\end{thebibliography}
\bibliographystyle{tmlr}


\clearpage

\appendix
\section*{Appendix}

In the appendix, we provide additional information as listed below:

\begin{itemize}
    \item \cref{sec:related} provides the related works.
    \item \cref{sec:dataset_metrics} provides the details of used datasets and metrics.
    \item \cref{sec:implementation} provides the implementation details.
    \item \cref{sec:ablation} provides extensive ablation study on the proposed \modelname.
    \item \cref{sec:more_vis} provides qualitative results on video reconstruction and video editing.
    \item \cref{sec:license} provides the licenses of the assets used.
    \item \cref{sec:broader_impact} discusses broader impact.
\end{itemize}

\section{Related Work}
\label{sec:related}
In this part, we introduce the literature related to the core part of \modelname (1st stage): neural rendering for scene representation (\cref{sec:nr_scene}), 3D Gaussian Spatting-based representation (\cref{sec:3dgs}), dynamic scene representation (\cref{sec:dynamic}), and spatial temporal decomposition in neural rendering (\cref{sec:sp}). Regarding to \modelname (2nd stage), we briefly review the literature on Text-to-Image editing (\cref{sec:t2i}), and Zero-shot Text-to-Video editing (\cref{sec:t2v}).

\subsection{Neural Rendering for Scene Representation}
\label{sec:nr_scene}
\vspace{-1mm}
The neural radiance field (NeRF) ~\citep{mildenhall2021nerf} is one of the advanced methodologies that represent scenes using implicit neural rendering with MLP. 
It has seen significant advancement recently, with major improvements in several key areas. Firstly, the speed of both training~\citep{Karnewar_2022} and inference~\citep{garbin2021fastnerf} has been substantially increased, allowing the current models to be trained and deployed within minutes~\citep{barron2021mipnerf, yu2021plenoctrees, reiser2021kilonerf, martinbrualla2021nerf}. 
Further advancements have addressed image artifacts, including motion blur~\citep{ma2022deblurnerf}, exposure and lens distortion~\citep{jeong2021selfcalibrating}, enhancing the quality and realism of rendered scenes. The necessity for precise camera calibrations is becoming less stringent, thanks to new methods that allow for local camera refinement~\citep{chng2022garf, lin2021barf}.
More recent research has also extended NeRF applications beyond traditional uses, exploring areas such as generalization~\citep{wang2022generalizable, chen2021mvsnerf}, semantic understanding~\citep{Zhi:etal:ICCV2021, kundu2022panoptic}, robotics~\citep{zhou2023nerf}, and 3D image style transfer~\citep{gu2021stylenerf}. These developments indicate a broadening scope and increasing versatility of NeRF models in capturing and rendering complex scenes.

\subsection{3D Gaussian Spatting-based Representation}
\label{sec:3dgs}
\vspace{-1mm}
Building upon the advancements in Neural Radiance Fields (NeRF)~\citep{kerbl20233d} for scene representation, another innovative approach that has gained attention is the utilization of 3D Gaussian distributions to model scenes. This methodology, often referred to as ``Gaussian Scene Representation'', leverages the mathematical properties of Gaussian distributions to efficiently capture the volumetric density and color information of 3D scenes.
Similar to the trend in NeRF, there have been several advancements in performance and efficiency. For example, some of them apply the module used in NeRF to 3D Gaussian-based view synthesis. 
~\citep{lee2024deblurring} leverages a compact MLP to adjust the covariance of 3D Gaussians, which can effectively reconstruct sharp details from blurry images and Mip-splatting~\citep{Yu2023MipSplatting} address alias issue in 3DGS as similar to ~\citep{barron2021mipnerf}. Besides the advancements in quality, ~\citep{fan2023lightgaussian} enhances the efficiency and compactness of 3D Gaussian Splatting for neural rendering by pruning insignificant Gaussians. 

\subsection{Handling Dynamic Scene Representation}
\label{sec:dynamic}
Despite the significant progress in NeRF and 3DGS for static scene representation, both approaches encounter substantial challenges when applied to dynamic scenes. The core limitation stems from their inherent design, which is primarily tailored for static environments, thus struggling to adapt to changes in scene geometry and appearance over time. To address these challenges, recent research endeavors have begun exploring extensions and alternatives to NeRF and 3D Gaussian Splatting that incorporate temporal information into the scene modeling process. For instance, 
deformation network has been widely used for encoding time dimension both in the framework of NeRF and 3DGS, 
~\citep{fan2023lightgaussian} and ~\citep{yang2023deformable3dgs} respectively. More developments for dynamic scene representation are introduced in NeRF. Robust-Dyn~\citep{liu2023robust} enhances the robustness of dynamic radiance field reconstruction by jointly estimating static and dynamic scene components along with camera parameters. 
NeRV~\cite{chen2021nerv} utilizes neural implicit representation to encode videos by simply fitting a neural network to input video frames.
Yet, despite their advancements, due to the fundamental limitation in the NeRF framework, these approaches continue to grapple with challenges related to achieving fine detail, maintaining temporal consistency across frames, and requiring extensive optimization times.
Driven by these challenges, the proposed \modelname seeks to investigate the application of 3DGS for dynamic scenes captured through monocular video, exploring its efficacy and potential improvements.

\subsection{Spatial Temporal Decomposition}
\label{sec:sp}
Our proposed \modelname decomposes the dynamic video spatially and temporally to represent original video effectively. Similarly, ProgressiveNeRF~\citep{meuleman2023localrf} proposes method of decomposing video into spatial and temporal components through progressively optimized local radiance fields. It represents a significant advancement in the field of robust view synthesis and large-scale scene reconstruction.
The primary distinction between our method and the ProgressiveNeRF approach lies in how we manage camera poses and frame integration across video clips. While ProgressiveNeRF dynamically updates camera poses and continues to append frames until the camera's pose extends to the boundary of a high-resolution, uncontracted space, our method treats each clip independently. We utilize overlapping frames solely to ensure consistency of 3D Gaussians corresponding to overlapping frames across neighboring clips, with each clip maintaining its unique camera poses. However, all clips can still be rendered onto the same image plane via a differentiable rendering with each distinctive 3D Gaussian and camera pose. Additionally, our process repeats until MC-COLMAP signals a `Success' \textit{status} marking a clear difference in the criteria used to determine the sufficiency of frames for each clip.
Furthermore, ProgressiveNeRF uses several dependencies (\eg, depth estimation and optical flow) and many losses for better temporal decomposition. However, we only need segmentation masks and simple reconstruction losses (L1 and SSIM) as in the original 3DGS paper.
We also provide extensive visual comparison with ProgressiveNeRF in this \href{https://anonymous-video-3dgs-progressivenerf.github.io/}{link}.

\subsection{Text-to-Image Editing}
\label{sec:t2i}
Text-to-image editing has significantly advanced with the development of generative models, particularly latent diffusion models ~\citep{rombach2022high}. These models have demonstrated remarkable capabilities in generating high-quality images guided by textual prompts. Methods such as DreamBooth~\citep{ruiz2022dreambooth} and Textual Inversion~\citep{gal2022textual} have achieved notable success by fine-tuning pre-trained models for specific editing tasks. Recent approaches, like Prompt-to-Prompt~\citep{hertz2022prompttoprompt} and DiffEdit~\citep{couairon2022diffedit}, leverage attention mechanisms within these models to enable localized and detailed image edits without extensive retraining. Techniques like SDEdit~\citep{meng2022sdedit} utilize stochastic differential equations to guide image synthesis, enabling precise control over the generated content. 
In addition to these, Blended Diffusion~\citep{Avrahami_2022_CVPR} presents a method to blend newly generated content seamlessly into existing images. This technique ensures that the synthesized additions are coherent with the original image context, thus maintaining a high level of realism.
Another notable approach is Paint by Word~\citep{bau2021paintbyword}, which introduces a framework where users can interactively edit images by providing textual descriptions for specific regions. This method utilizes a combination of mask prediction and image generation to accurately reflect the user's intent. 
Despite significant advancements in text-to-image editing, the progress in text-to-video editing technology has been relatively slow.

\subsection{Zero-shot Text-to-Video Editing}
\label{sec:t2v}
Zero-shot text-to-video editing leverages pre-trained text-to-image models to edit videos without requiring extensive retraining. This approach aims to overcome the limitations of traditional video editing methods, such as video-specific training~\citep{ho2022video} and atlas learning~\citep{kasten2021layered}, which are often time-consuming and require significant manual effort. Recent advancements in this domain have introduced various techniques to achieve high-quality and temporally consistent video edits.
For instance, RAVE~\citep{2312.04524} employs a noise shuffling strategy to enhance temporal consistency by leveraging spatio-temporal interactions between frames. This method integrates spatial guidance through ControlNet~\citep{zhang2023adding} and operates efficiently in terms of memory requirements, making it suitable for editing longer videos. RAVE's approach significantly reduces processing time compared to existing methods, achieving roughly 25\% faster editing rate while maintaining high visual quality.
Other notable zero-shot methods include Pix2Video~\citep{ceylan2023pix2video} and FateZero~\citep{qi2023fatezero}, which utilize sparse-causal attention and attention feature preservation, respectively, to maintain motion and structural consistency across video frames. Text2Video-Zero~\citep{text2video-zero} synthesizes and edits videos by integrating cross-frame attention and controlling the fidelity to structure of original video with image guidance scale.
TokenFlow~\citep{tokenflow2023} enforces consistency by propagating diffusion features across frames based on inter-frame correspondences in zero-shot manner.
These techniques collectively demonstrate the potential of zero-shot video editing in producing temporally coherent and visually appealing videos without the need for extensive training on video-specific datasets.
Despite the recognized improvements in efficiency and editing capabilities offered by zero-shot video editors, they still suffer from subpar temporal consistency and overall video editing quality due to a lack of per-scene understanding. This limitation motivates us to design a plug-and-play refiner for zero-shot video editors (we select representative three editors: Text2Video-Zero / TokenFlow / RAVE), aimed at enhancing per-scene understanding and improving the quality of the edited videos.





\section{Datasets and Metrics}
\label{sec:dataset_metrics}

\subsection{Datasets}

 \textbf{Video Reconstruction}\quad
To assess the quality of video reconstruction in real-world monocular video scenes, we curated a dataset comprising 28 representative videos from the DAVIS dataset~\citep{Pont-Tuset_arXiv_2017}, each with a resolution of 480p. These videos feature a varied array of foreground moving objects, encompassing humans, vehicles, and animals, together with cluttered backgrounds. Belows are the key features of DAVIS dataset:

\begin{itemize}
    \item Resolution: videos are available in two resolutions, 480p and 1080p. Following prior works, we utilized 480p.
    \item Each video has a different number of frames ranging from 40 to 80
    \item FPS: 24
    \item Video sequences: the dataset includes 50 video sequences. Among them, we selected 28 videos that feature a varied array of foreground-moving objects, encompassing humans, vehicles, and animals.

\end{itemize}


 \textbf{Video Editing}\quad
For video editing, we adopt the datasets and text prompts used in the CVPR 2023 workshop, LOVEU Text-Guided Video Editing (TGVE) challenge~\citep{wu2023cvpr}.
They curated a dataset consisting of 76 videos (480p) in-the-wild from DAVIS, Youtube, and Videvo Videos. Each video is edited according to 4 different types of text prompts acquired from BLIP-2~\citep{li2023blip}, including: style change, object change, background change, and multiple change.
As noted in the main paper, we tested our framework in 58 videos with 4 editing prompts, totaling 232 video editing scenarios. Below are key details.

\begin{itemize}
    \item Resolution: 480x480
    \item Frame number: DAVIS: 32 / Youtube: 128 / Videvo: 32
    \item FPS: 24

\end{itemize}










\vspace{-1mm}

\subsection{Evaluation Metrics}
\label{sec:metrics}

\vspace{-1mm}
\textbf{Video Reconstruction}\quad
To evaluate video reconstruction quality, two principal metrics are utilized: Peak Signal-to-Noise Ratio (PSNR) and Structural Similarity Index Measure (SSIM). These metrics collectively appraise the fidelity of the reconstructed video in comparison to the original. Additionally, efficiency is gauged through the metric of Training Time, which quantifies the required training time, thus indicating the computational efficiency of the reconstruction algorithm.


\textbf{Video Editing}\quad
To assess video editing quality, we utilize WarpSSIM, which computes the mean SSIM score between the edited video warped by the optical flow~\citep{teed2020raft} (derived from the source video) and corresponding original edited video.
This metric offers valuable insight into the temporal consistency of post-editing. Furthermore, we employ $Q_{edit}$~\citep{2312.04524}, a comprehensive video editing metric that combines the WarpSSIM score with the CLIPScore\_text, providing a multiplicative assessment of the overall video editing performance.

\section{Implementation Details} 
\label{sec:implementation}
Our framework is based on Pytorch~\citep{paszke2019pytorch} and 3D Gaussian Splatting~\citep{kerbl20233d}, while adopting depth incorporated differentiable Gaussian rasterization~\citep{yang2023deformable3dgs}. During training, two sets of 3D Gaussians (both Frg-3DGS and Bkg-3DGS) in each clip are sequentially optimized with three different total numbers of training iterations: a total of 3k, 5k and 10k iterations, each targeting a distinct time-accuracy trade-off (\eg, 10k iterations for the best quality but the slowest training time).
We employ the same training hyper-parameters (\eg, learning rate, deformation network setting) as specified by~\citep{yang2023deformable3dgs}. 
With optimized 3D Gaussians for each clip, video reconstruction proceeds through sequential differential rendering~\citep{kerbl20233d} of the clips.
For video editing, 3D Gaussians optimized for original videos undergo fine-tuning for 1k iterations to update the spherical harmonic coefficients (SH) and opacity ($\sigma$), guided by the initial edited videos (obtained from an off-the-shelf zero-shot video editor).
Each experiment uses a single A100 GPU.

\begin{table*}
\centering
\resizebox{1.0\linewidth}{!}{
\begin{tabular}{c|c|cccc|cc|cc}
\toprule
\toprule
\multirow{3}{*}{COLMAP} & Type                      & \multicolumn{4}{c|}{COLMAP}                                                                            & \multicolumn{2}{c|}{Masked COLMAP}            & \multicolumn{2}{c}{MC COLMAP}                \\ \cline{2-10} 
                        &  success rate   & \multicolumn{4}{c|}{\graycell{20/28}}                                                                             & \multicolumn{2}{c|}{\yellowcell{25/28}}                    & \multicolumn{2}{c}{\redcell{28/28}}                    \\
                        & processing time           & \multicolumn{4}{c|}{\graycell{12m30s}}                                                                              & \multicolumn{2}{c|}{\yellowcell{2m10s}}                     & \multicolumn{2}{c}{\redcell{2m3s}}                     \\ \hline
\multirow{3}{*}{3DGS}   & Type                      & \multicolumn{1}{c|}{Deformable-3DGS} & \multicolumn{3}{c|}{Frg-3DGS}                                 & \multicolumn{2}{c|}{Frg-3DGS + Bkg-3DGS} & \multicolumn{2}{c}{Frg-3DGS + Bkg-3DGS} \\ \cline{2-10} 
                        & iteration / training time & \multicolumn{1}{c|}{40k (50m)}           & \multicolumn{1}{c}{\redcell{10k (4m)} } & \multicolumn{1}{c}{30k (15m)} & 60k (34m) & \yellowcell{10k (8m)}        & \multicolumn{1}{c|}{30k (34m)}       & \multicolumn{1}{c}{\graycell{3k (11m)}}          & \multicolumn{1}{c}{5k (21m)}         \\
                        & PSNR               & \multicolumn{1}{c|}{30.6}                & 32.9      &   35.3    &  37.3     &  35.1          & \yellowcell{38.0}            &  \graycell{37.6}          & \redcell{41.2}         \\

\bottomrule
\bottomrule
\end{tabular}
}
\caption{Ablation study on MC-COLMAP.
The colored cells indicate the \sethlcolor{lightred}\hl{best}, \sethlcolor{lightyellow}\hl{second best}, and \sethlcolor{lightgray}\hl{third best} performances (best viewed in color). Our \colmap not only successfully generates 3D points for all 28 videos, but also enables the high reconstruction quality.}
\label{tab:mccolmap}
\end{table*}
\begin{table*}
\centering
\resizebox{0.5\linewidth}{!}{
    \begin{tabular}{l|c|c|c|c}
        \toprule
        \toprule
        Scene & blackswan & boat & drift-turn & horsejump-high \\ \hline
        Clip \#1 & 17(1) & 11(1) & 10(1) & 10(1) \\ \hline
        Clip \#2 & 11(1) & 13(1) & 11(1) & 11(1) \\ \hline
        Clip \#3 & 11(1) & 17(1) & 11(1) & 11(1) \\ \hline
        Clip \#4 & 11(1) & 20(5) & 11(1) & 11(1) \\ \hline
        Clip \#5 & 4 & 22 & 11(1) & 11 \\ \hline
        Clip \#6 & - & - & 11(1) & - \\ \hline
        Clip \#7 & - & - & 8 & - \\ \bottomrule \bottomrule
    \end{tabular}

    }
    \caption{Analysis on statistics of MC-COLMAP. We show the number of frames for each clip in different scenes. The number in the parentheses refers to the number of overlapping frames between neighboring clips.}
    \label{tab:clip_table}

\end{table*}



\section{Ablation Study}
\label{sec:ablation}

\subsection{\modelname (1st Stage)}
\label{sec:ablation_first}
\textbf{Analysis on MC-COLMAP}\quad
We present a comprehensive analysis of MC-COLMAP.
\cref{tab:mccolmap} compares two baselines: conventional COLMAP, which processes full video frames, and Masked COLMAP, which employs only spatial decomposition.
Conventional COLMAP encounters more failure case (8 out of 28 videos fail to obtain 3D points) and exhibits slower processing time (12m30s).
Among the successful cases (20 videos),
reconstruction using the state-of-the-art Deformable-3DGS~\citep{yang2023deformable3dgs} performs suboptimally in both training time and PSNR compared to our Frg-3DGS, powered by multi-resolution hash encoding-based deformation network.
It is worth noting that this study utilizes only one set of 3D Gaussians, without spatial decomposition for foreground and background points.
Masked COLMAP demonstrates improved success rate (only 3 out of 28 videos fail to obtain 3D points) and enhances the processing speed by mitigating the effects of cluttered backgrounds through spatial decomposition.
Additionally, reconstruction quality, as measured by PSNR, is improved.
Finally, our proposed MC-COLMAP, which incorporates both spatial and temporal decomposition, successfully captures 3D points from all 28 videos.
Furthermore, it enables the best reconstruction quality of 41.2 PSNR with 5k training iterations, significantly outperforming the other settings.

\textbf{MC-COLMAP Statistics}\quad
As shown in~\cref{tab:clip_table}, to enhance comprehension of MC-COLMAP, we present statistics that illustrate the distribution of clip length across four representative scenes from the challenging DAVIS dataset.
We have selected a default value of k=10 for our experiments and found out that altering k to either 5 or 15 does not markedly impact our results.
We synchronize different 3DGS sets across video clips through the use of overlapping frames, which are indicated in the above table as numbers within parentheses. Initially, we designate the count of overlapping frames to be one, although this number can be adjusted based on the number of frames added to each clip. It also shows behaviors for slow or fast motion in video.
For example, the scenes ``blackswan'' and ``boat'' feature comparatively static and slow movements, whereas ``drift-turn'' and ``horsejump-high'' are characterized by highly dynamic actions. Each scene begins with an initial set of 10 frames for MC-COLMAP processing. It is noted that fast-moving scenes are typically decomposed into 10 or 11 frames. In contrast, scenes with slower motion demand a greater number of frames to accurately extract 3D points for MC-COLMAP analysis.

\begin{table}[t!]
  \centering
  \scalebox{0.95}{
  \begin{tabular}{c|c|c}
      \toprule
      \toprule
      \diagbox{Method}{Metrics}& PSNR & SSIM \\ \hline
      Frg-3DGS (w/ fore) & 35.1 & 0.923 \\
      Frg-3DGS (w/ rdn) & 36.5 & 0.951 \\ 
      Frg-3DGS (w/ fore+rdn) & 37.5 & 0.957 \\ \hline
      Frg-3DGS + Bkg-3DGS & 41.4 & 0.987 \\ 
      \bottomrule
      \bottomrule
    \end{tabular}
    }
    \vspace{2.5mm}
    \caption{Analysis on Frg-3DGS and Bkg-3DGS. `fore' and `rnd' denote the foreground and random 3D points, respectively. Utilizing a single set of Frg-3DGS can benefit from both types of 3D points. However,  performance can be enhanced further by employing two sets of Frg-3DGS and Bkg-3DGS, tailored specifically for foreground and background random 3D points, respectively.
     }
    \label{tab:dual}
    \vspace{-4mm}
\end{table}

\textbf{Ablations on Segmentation Modules}\quad
we additionally include the ablation study on using different video segmentation modules. There are several open-vocabulary video object segmentation models available, and we have chosen the most representative models for our experiments, including DEVA model with SAM, and Mobile-SAM. We experimented with three settings: (a) DEVA with SAM and user prompts (i.e., user needs to provide the object class names), (b) DEVA with SAM and VIPSeg foreground categories (i.e., we use the foreground categories defined by VIPSeg dataset without any user prompts), and (c) DEVA with Mobile-SAM and VIPSeg foreground categories. As shown in this \href{https://anonymous-video-3dgs-seg-ablation.github.io}{video}, our Video-3DGS is robust to the change of segmentation modules. Furthermore, it can also prove that our MC-COLMAP effectively reconstructs foreground moving objects regardless of their predefined categories, thereby achieving comparable reconstruction capabilities.

\begin{table}[t!]
    \centering
    \begin{minipage}{0.5\textwidth}
        \centering
        \resizebox{0.97\textwidth}{!}{ 
        \begin{tabular}{lcccc}
            \toprule
            \textbf{DAVIS (480p)} & \textbf{PSNR} & \textbf{SSIM} & \textbf{WarpSSIM} & \textbf{Qedit} \\
            \midrule
             CoDeF & 28.7 & 0.904 & 0.926 & 23.7 \\
             Video-3DGS (3k) & 40.8 & 0.986 & - & - \\
             CoDeF+Video-3DGS & - & - & 0.901 & 23.2 \\
             RAVE & - & - & 0.872 & 23.4 \\
             RAVE+Video-3DGS & - & - & 0.908 & 24.1 \\
             RAVE+Video-3DGS (RE) & - & - & 0.913 & 24.8 \\
            \bottomrule
        \end{tabular}
        }
    \end{minipage}%
    \hfill
    \begin{minipage}{0.5\textwidth}
        \centering
        \resizebox{0.97\textwidth}{!}{ 
        \begin{tabular}{lcccc}
            \toprule
            \textbf{Youtube} & \textbf{PSNR} & \textbf{SSIM} & \textbf{WarpSSIM} & \textbf{Qedit} \\
            \midrule
             CoDeF & 25.5 & 0.795 & 0.925 & 21.8 \\
             Video-3DGS (3k) & 40.8 & 0.986 & - & - \\
             CoDeF+Video-3DGS & - & - & 0.924 & 21.2 \\
             TokenFlow & - & - & 0.848 & 22.1 \\
             TokenFlow+Video-3DGS & - & - & 0.923 & 23.1 \\
             TokenFlow+Video-3DGS (RE) & - & - & 0.923 & 24.2 \\
            \bottomrule
        \end{tabular}
        }
    \end{minipage}
    \caption{Comparison of the proposed \modelname method with the CoDEF~\cite{ouyang2023codef} framework in the context of video editing tasks. The results demonstrate that \modelname exhibits strong compatibility with zero-shot video editors, yielding superior or comparable performance to CoDEF across several video editing benchmarks.}
    \label{tab:codef_comp}
\end{table}

\textbf{Analysis on Frg-3DGS and Bkg-3DGS}\quad
In \cref{tab:dual}, we demonstrate the necessity of employing two sets of Frg-3DGS and Bkg-3DGS for video reconstruction.
We begin with a single set of Frg-3DGS that leverages only the foreground 3D points (denoted as Frg-3DGS w/ fore), resulting in 35.1 PSNR and 0.923 SSIM.
By initializing the single set of Frg-3DGS with the spherical-shaped random points (denoted as Frg-3DGS w/ rdn), the performance improves to 36.5 PSNR and 0.951 SSIM.
The inclusion of random 3D points uniformly captures both foreground and background points, thus enhancing performance.
Furthermore, employing both foreground 3D points and spherical-shaped random 3D points (Frg-3DGS w/ fore+rdn) yields further improvement.
Finally, utilizing two sets of Gaussians, Frg-3DGS and Bkg-3DGS, each specific to foreground and background 3D points respectively, significantly enhances performance to 41.4 PSNR and 0.987 SSIM.

\textbf{2D Learnable Parameter for Merging Foreground and Background Views}\quad
We visualized the 2D learnable parameter (\ie, $\alpha$ in~\cref{eqn:merge}) in different video scenarios in the following \href{https://anonymous-video-3dgs-2d-learnable-mask.github.io/}{link}. As shown in the videos, the learned $\alpha$ focuses on the foreground regions, while the learned (1 - $\alpha$) focuses on the background regions.

\textbf{Model Size}\quad
The overall size of the model is determined by both the number of clips and the size of the Frg-3DGS and Bkg-3DGS for each clip. 
For example, consider the ``blackswan'' scene. the Video-3DGS (3k iteration) necessitates, on average, a 15MB Frg-3DGS and a 60MB Bkg-3DGS per clip. Consequently, the cumulative model size amounts to 375MB. In contrast, the conventional 3DGS (30k iteration) employs a singular, substantial 3DGS model, approximately 300MB in size, which is a result of the densification technique employed during long iterations. Despite the relatively big model size required for 3DGS, since it represents and stores the scene in an explicit manner (point cloud), we can take advantage of the efficient training and inference capability of 3DGS in videos.
It should be noted that the aggregate size of the Video-3DGS model escalates with an increasing number of clips. Therefore, a pivotal objective for future research is to minimize the overall size requisite for Video-3DGS implementation. One of the naive ways of achieving this is opting for a larger k value so that we can effectively reduce the final model size, as this leads to a decreased number of clips.

\subsection{\modelname (2nd Stage)}

\textbf{Comparison With NeRF-based Video Editing}\quad
\label{sec:codef_compare}
In addition to \cref{tab:edit_quan} where we show the compatibility of \modelname with zero-shot video editors, we compare with a training-based video editor that utilizes NeRF~\cite{mildenhall2021nerf}
CoDEF~\cite{ouyang2023codef} as in \cref{tab:codef_comp}.  We can observe that \modelname with zero-shot video editors (e.g., RAVE or TokenFlow) show comparable or better results than “CoDeF+Video-3DGS” in terms of both reconstruction and editing ability. 
Qualitative Results with CoDEF can be found in this \href{https://anonymous-video-3dgs-codef-exps.github.io}{link}.
As shown in the qualitative results, CoDEF struggles to preserve the content and structure of the original video due to its application of ControlNet on a canonical image. In contrast, our Video-3DGS effectively maintains the original content and structure.

\begin{table}[t!]
  \centering
  \scalebox{0.72}{
  \begin{tabular}{c|c|c}
      \toprule
      \toprule
      \diagbox{Method}{Metrics} & WarpSSIM↑ & $Q_{edit}$↑ \\ \hline
      \multicolumn{1}{l|}{UniEdit~\citep{bai2024uniedit}} & 0.779 & 19.95 \\
      \multicolumn{1}{l|}{+\modelname}& 0.830 & 21.15 \\ 
      \bottomrule
      \multicolumn{1}{l|}{AnyV2V~\citep{ku2024anyv2v}} & 0.737 & 19.85 \\ 
      \multicolumn{1}{l|}{+\modelname} & 0.836 & 22.25 \\ 
      \bottomrule
    \end{tabular}
    }
    \vspace{2.5mm}
    \caption{Additional results on compatibility of \modelname with video editors using video generation models. Our \modelname shows consistent compatibility with video editors using video generation model by increasing the editing quality. We used 16 DAVIS videos with 32 frames and initially edited them with style-change prompts from LOVEU~\citep{wu2023cvpr} benchmark.
     }
    \label{tab:compat}
    \vspace{-4mm}
\end{table}

\vspace{2mm}

\textbf{Comparison With Editors Using Video Generation Models}\quad
\label{sec:compat}
Recently, a new trend in video editing has emerged by leveraging video generation models. These approaches can be broadly categorized into two types: those that use image-to-video (I2V) priors and those that employ text-to-video (T2V) priors. To illustrate, we selected a representative method from each category—UniEdit~\citep{bai2024uniedit} (T2V-based) and AnyV2V~\citep{ku2024anyv2v} (I2V-based).
Due to UniEdit’s high memory requirements, we divided DAVIS videos (32 frames each) into four chunks and applied the LaVie~\citep{wang2024lavie} T2V model to each chunk as a prior for video editing. For AnyV2V, we edited the first frame using Instruct-pix2pix~\citep{brooks2022instructpix2pix} and then propagated the edits to the remaining frames using its I2V model.
After obtaining the initial outputs from each methods, we refined the results with our \modelname. As shown in \cref{tab:compat}, \modelname demonstrates strong compatibility with recent methods that use video generation priors, particularly when processing DAVIS videos with style-change prompts from the LOVEU benchmark.

\textbf{Time Cost Analysis}\quad
\label{sec:time_cost}
Similar to \cref{tab:recon_sum_quan}, we also explore the time efficiency of video editing when incorporating \modelname atop existing zero-shot video editing frameworks in this section in \cref{tab:timecost}. This assessment spans three critical phases: optimizing Frg-3DGS and Bkg-3DGS with the initial video, updating $SH$ and $\sigma$ for the initially edited video, and rendering to produce a refined edit. We can demonstrate that integrating \modelname as an enhancement module consistently requires approximately 6 minutes, irrespective of the underlying editing techniques, affirming its efficiency as a reasonable investment in processing time.

\textbf{Analysis on Recursive and Ensembled Refinement}\quad
\label{sec:re_vid3dgs} 
After observing the effectiveness of \modelname (RE) in \cref{tab:edit_quan}, we conducted an ablation study on its components, as shown in \cref{tab:re_analysis}. Each component individually improves performance compared to a single-phase refiner. The `Recursive' component increases WarpSSIM by 1.5\% and $Q_{edit}$ by 0.4\%. Similarly, the `Ensembled' component enhances the refiner, resulting in a 7\% increase in WarpSSIM and a 0.4\% increase in $Q_{edit}$. Our \modelname (RE), which combines both recursive and ensembled strategies, achieves the highest scores in both WarpSSIM (\textbf{0.899}) and $Q_{edit}$ (\textbf{22.3}).


\subsection{Relationship between \modelname (1st Stage) and \modelname (2nd Stage)}

We examine the correlation between video reconstruction and editing quality in~\cref{tab:rel_rec_edit}.
We introduce two versions of Video-3DGS: model A, where all clips utilize the same deformation field (thus reducing the representation capacity), and model B, our default setting. As depicted in the table, model B achieves superior reconstruction quality compared to model A, resulting in better performance across all three off-the-shelf video editors. We also conducted qualitative comparison with 3DGS~\citep{kerbl20233d} to further prove the positive relationship between video reconstruction and video editing in this \href{https://anonymous-video-3dgs-recon-edit.github.io/}{link}.






\begin{table*}
\centering
\resizebox{0.8\linewidth}{!}{
\begin{tabular}{c|cc|cc|cc}
\toprule
\toprule
             & Text2Video-Zero & + Video-3DGS & TokenFlow & + Video-3DGS & RAVE & +Video-3DGS \\ \hline
Editing time & 1m11             & +6m7s         & 2m50s      & +6m7s         & 7m20s & +6m5s        \\
\bottomrule
\bottomrule
\end{tabular}
}
\caption{Ablation study on time cost of \modelname (`style change' in DAVIS).
}
\label{tab:timecost}
\end{table*}

\begin{table}[t!]
\centering
\resizebox{0.65\linewidth}{!}{
\begin{tabular}{l||cc||cc|cc}
\toprule
\toprule
 \multirow{2}{*}{\diagbox{Dataset}{Method}} & \multicolumn{2}{c||}{Text2Vid-Zero} & \multicolumn{2}{c|}{+\modelname} & \multirow{2}{*}{WarpSSIM↑} & \multirow{2}{*}{$Q_{edit}$↑}   \\
\cmidrule(lr){2-3} \cmidrule(lr){4-5} 
 & WarpSSIM↑ & $Q_{edit}$↑ &  Recursive & Ensembled &  &   \\

\midrule
 \multirow{4}{*}{DAVIS} & \multirow{4}{*}{0.691} & \multirow{4}{*}{20.1} &  &  & 0.827 & 21.0\\

 &   &  & \checkmark &  & 0.842 & 21.4\\

 &   &  &  & \checkmark & 0.897 & 21.4 \\

 &   &  & \checkmark & \checkmark & \textbf{0.899} & \textbf{22.3} \\

\bottomrule
\bottomrule

\end{tabular}
}
\vspace{2mm}
\caption{Analysis on the components of recursive and ensembled \modelname. 
}
\label{tab:re_analysis}
\end{table}


\begin{table*}[t!]
\vspace{-1.5mm}
\centering
\scalebox{0.53}{
\begin{tabular}{c|ccc||c|ccc||c|ccc}
\toprule
\toprule
Text2Video-Zero               & \multicolumn{3}{c||}{+Video-3DGS}                       & TokenFlow                     & \multicolumn{3}{c||}{+Video-3DGS}                       & RAVE                          & \multicolumn{3}{c}{+Video-3DGS}                        \\ \hline
Editing                       & \multicolumn{2}{c|}{Reconstruction} & Editing          & Editing                       & \multicolumn{2}{c|}{Reconstruction} & Editing          & Editing                       & \multicolumn{2}{c|}{Reconstruction} & Editing          \\
WarpSSIM / Qedit              & Model  & \multicolumn{1}{l|}{PSNR}  & WarpSSIM / Qedit & WarpSSIM / Qedit              & Model  & \multicolumn{1}{l|}{PSNR}  & WarpSSIM / Qedit & WarpSSIM / Qedit              & Model  & \multicolumn{1}{l|}{PSNR}  & WarpSSIM / Qedit \\ \hline
\multirow{2}{*}{0.791 / 21.1} & A      & \multicolumn{1}{l|}{31.1}  & 0.842 / 21.7     & \multirow{2}{*}{0.869 / 23.2} & A      & \multicolumn{1}{l|}{31.1}  & 0.884 / 22.9     & \multirow{2}{*}{0.874 / 24.8} & A      & \multicolumn{1}{l|}{31.1}  & 0.871 / 23.3     \\ \cline{2-4} \cline{6-8} \cline{10-12} 
                              & B      & \multicolumn{1}{l|}{41.3}  & 0.841 / 22.8     &                               & B      & \multicolumn{1}{l|}{41.3}  & 0.916 / 24.4     &                               & B      & \multicolumn{1}{l|}{41.3}  & 0.905 / 25.6     \\
\bottomrule
\bottomrule
\end{tabular}}
\caption{The relationship between video reconstruction and video editing on DAVIS.
Across all three off-the-shelf video editors (Text2Video-Zero, TokenFlow, and RAVE), model B (with better reconstruction ability) yields better video editing results than model A. }
\label{tab:rel_rec_edit}
\end{table*}

\section{Qualitative Results}
\label{sec:more_vis}
In this section, we provide more qualitative results for both video reconstruction (\cref{sec:more_vis_recon}) and video editing (\cref{sec:more_vis_edit}).

\subsection{Video Reconstruction}
\label{sec:more_vis_recon}
\begin{figure*}[!t]
\begin{center}
\includegraphics[width=0.7\linewidth]{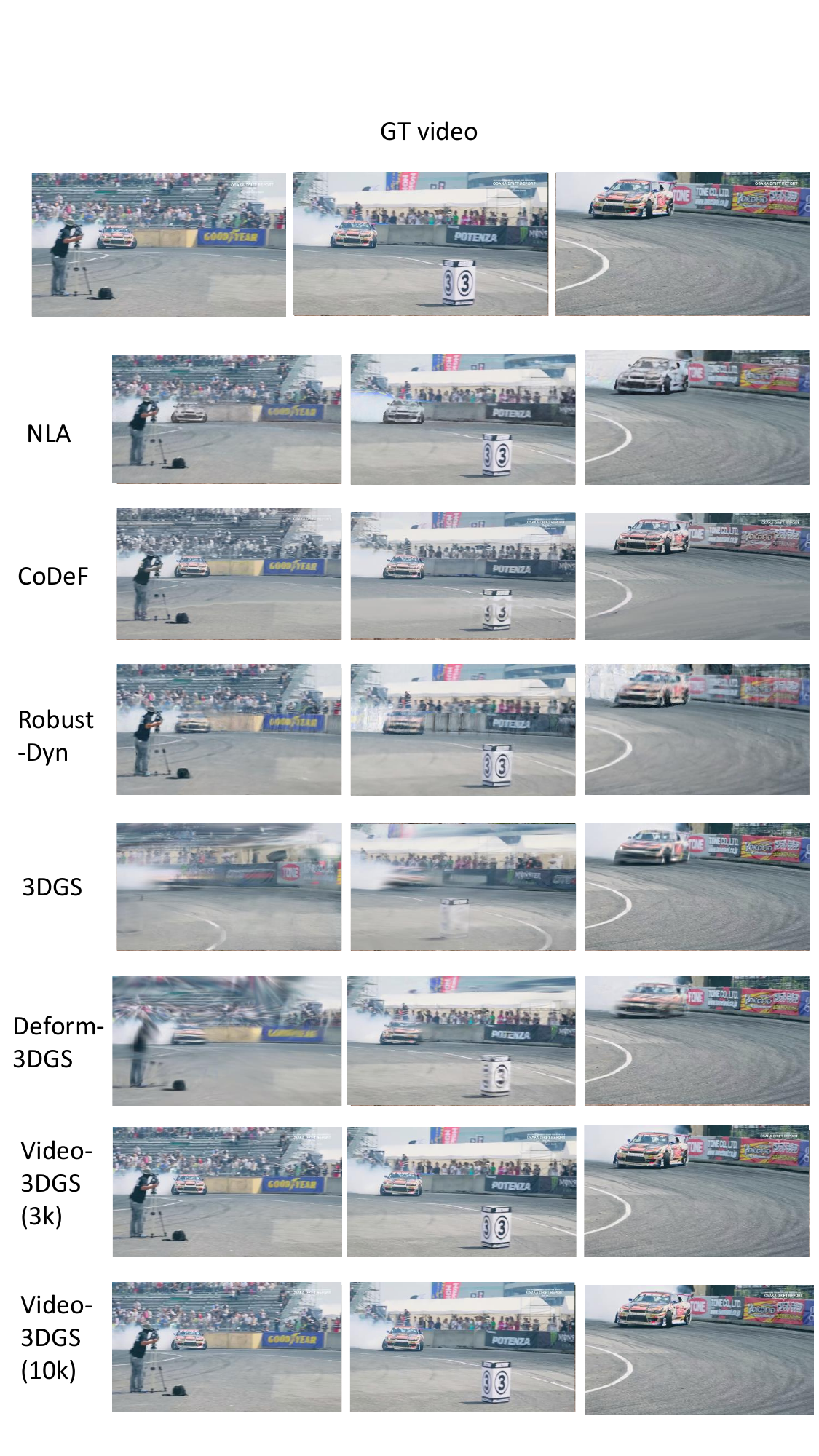}
\vspace{-4mm}
\caption{Qualitative results for video reconstruction.} 
\label{fig:recon1}
\end{center}
\end{figure*}

\begin{figure*}[!t]
\begin{center}
\includegraphics[width=0.7\linewidth]{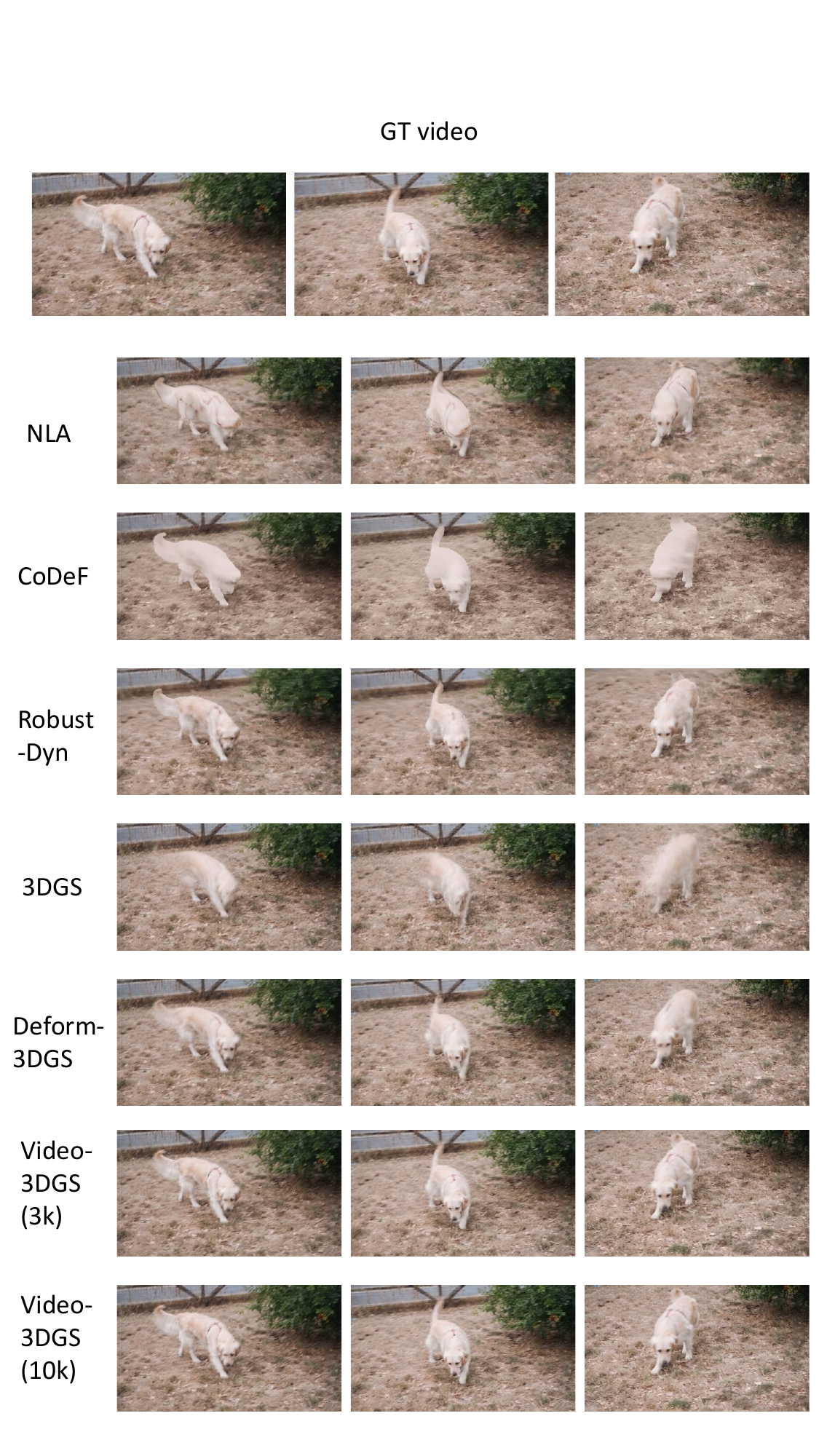}
\vspace{-4mm}
\caption{Qualitative results for video reconstruction.} 
\label{fig:recon2}
\end{center}
\end{figure*}

\begin{figure*}[!t]
\begin{center}
\includegraphics[width=0.7\linewidth]{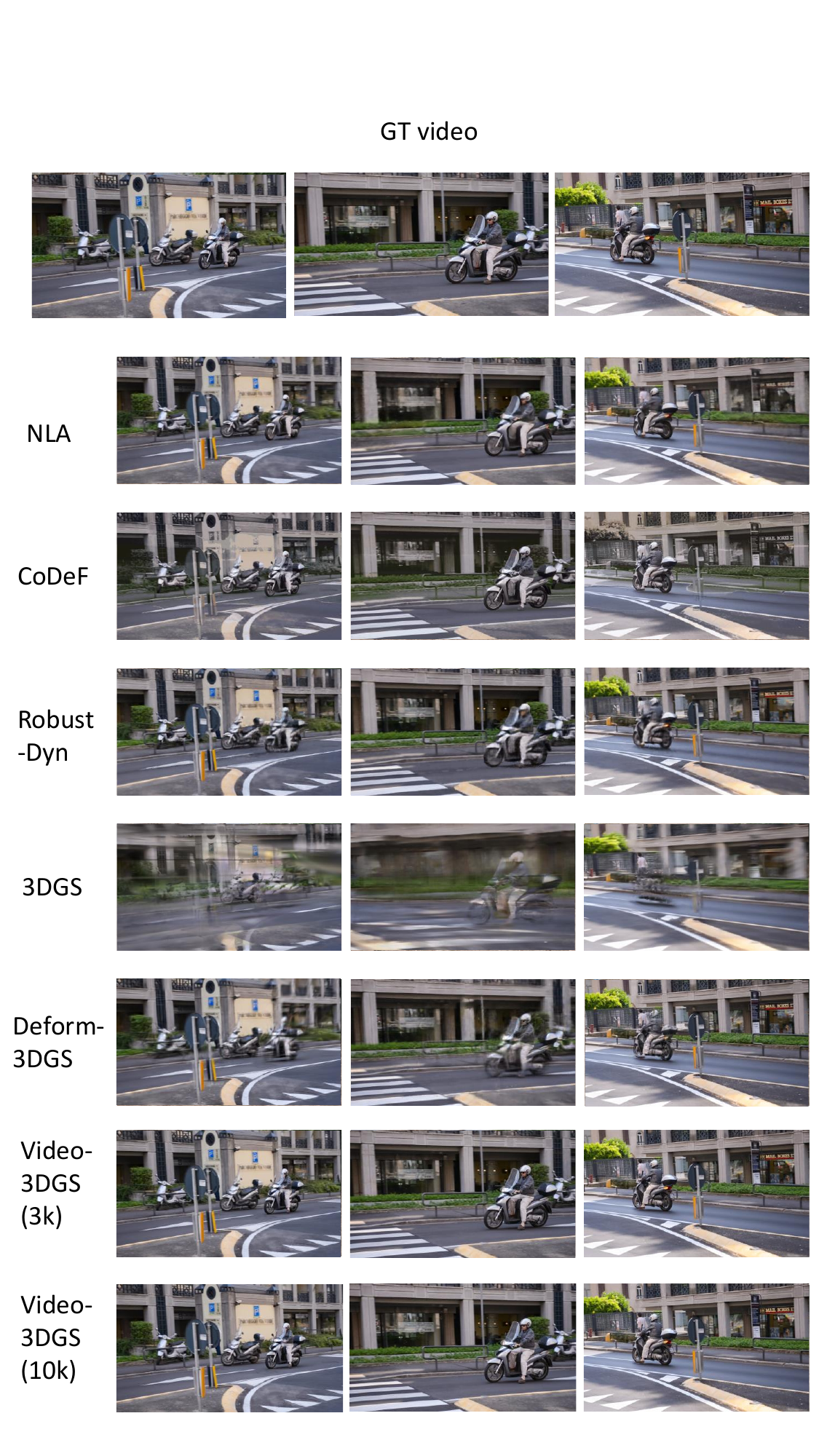}
\vspace{-4mm}
\caption{Qualitative results for video reconstruction.} 
\label{fig:recon3}
\end{center}
\end{figure*}

For video reconstruction, we visualize qualitative results in~\cref{fig:recon1}, \cref{fig:recon2}, and~\cref{fig:recon3}. As shown in the figures, the proposed \modelname consistently demonstrates higher reconstruction quality.
Video comparisons with 3DGS~\citep{kerbl20233d} and Deformable-3DGS~\citep{yang2023deformable3dgs} are provided in the following \href{https://anonymous-video-3dgs-additional-recon.github.io/}{link}

\subsubsection{Spatial Decomposition}
We also provide visual analysis on the spatial decomposition in the following video \href{https://anonymous-video-3dgs-spatial.github.io/}{link}. Our method successfully extracts the corresponding 3D points of moving objects using MC-COLMAP. We also found that the spatial decomposition can work well in a scene with multiple foreground objects. 

\subsection{Video Editing}
\label{sec:more_vis_edit}

For video editing, we provide the visualization results for the `Single-phase Refiner' and `Recursive and Ensembled Refiner'.

\subsubsection{Single-phase Refiner}
\begin{figure*}[!t]
\begin{center}
\includegraphics[width=0.9\linewidth]{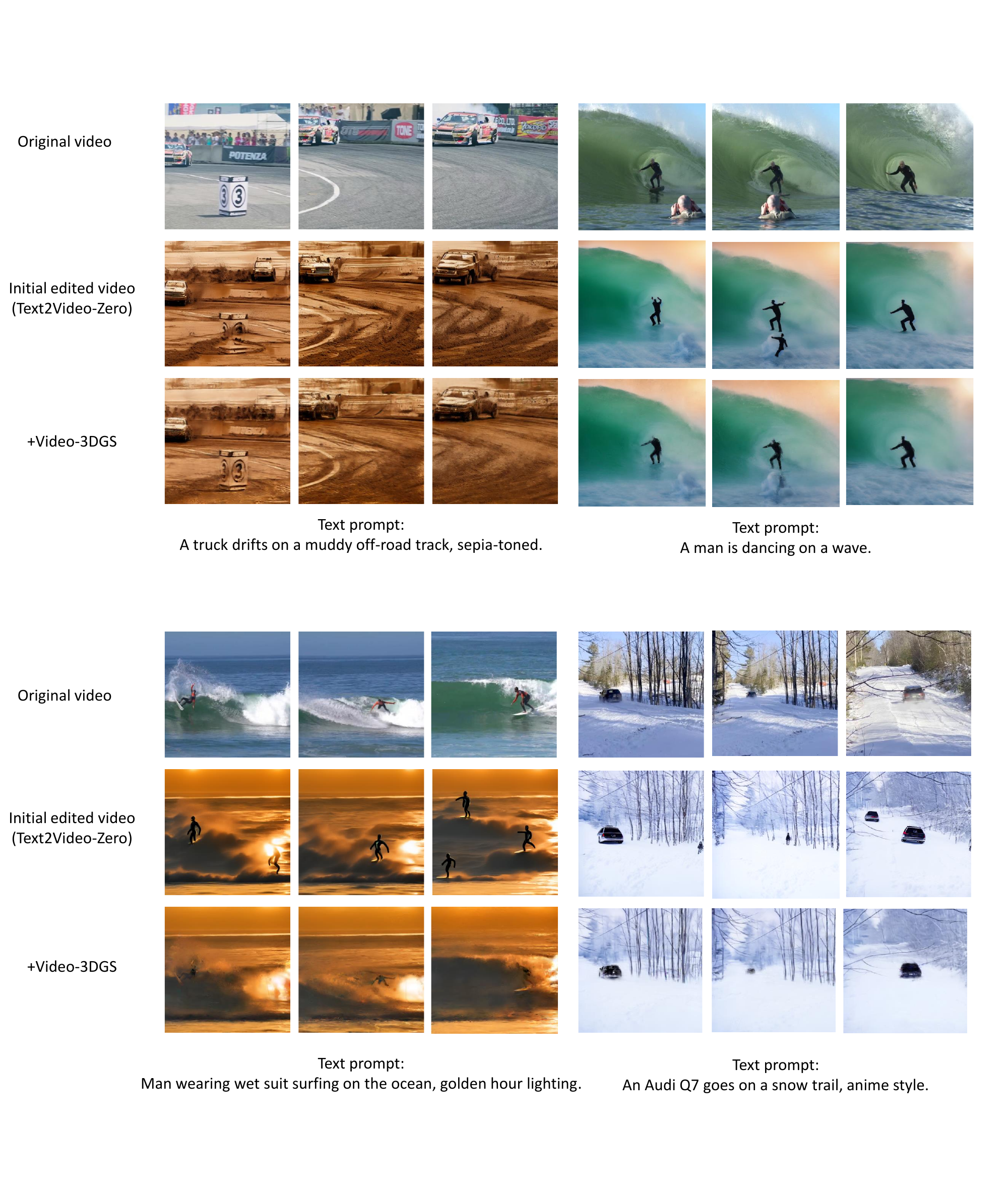}
\vspace{-4mm}
\caption{Qualitative results of single stage refiner for video editing by adopting Text2Video-Zero.
} 
\label{fig:edit1}
\end{center}
\end{figure*}

\begin{figure*}[!t]
\begin{center}
\includegraphics[width=0.9\linewidth]{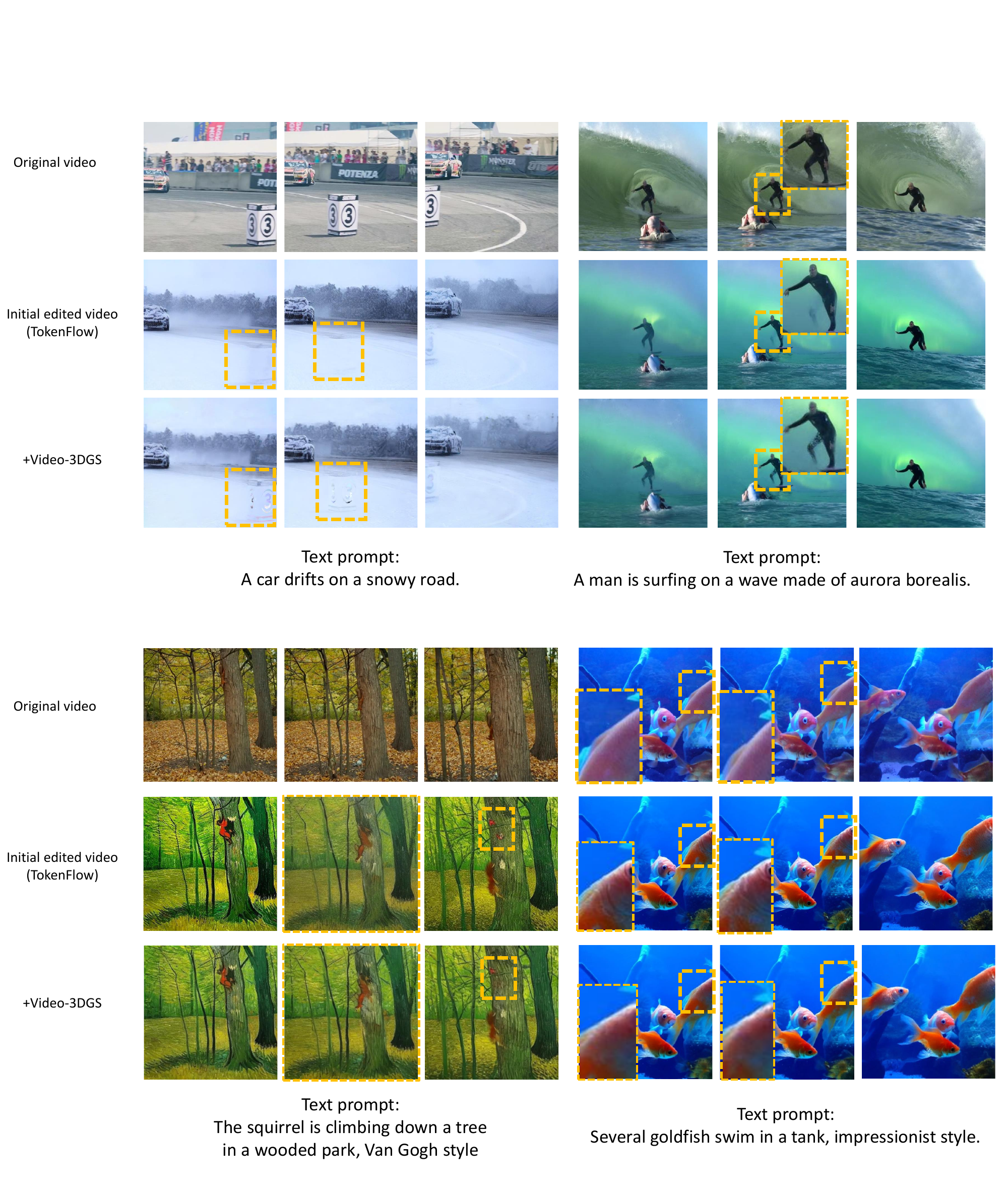}
\vspace{-4mm}
\caption{Qualitative results of single stage refiner video editing by adopting TokenFlow.} 
\label{fig:edit2}
\end{center}
\vspace{-5mm}
\end{figure*}

\begin{figure*}[!t]
\begin{center}
\includegraphics[width=0.9\linewidth]{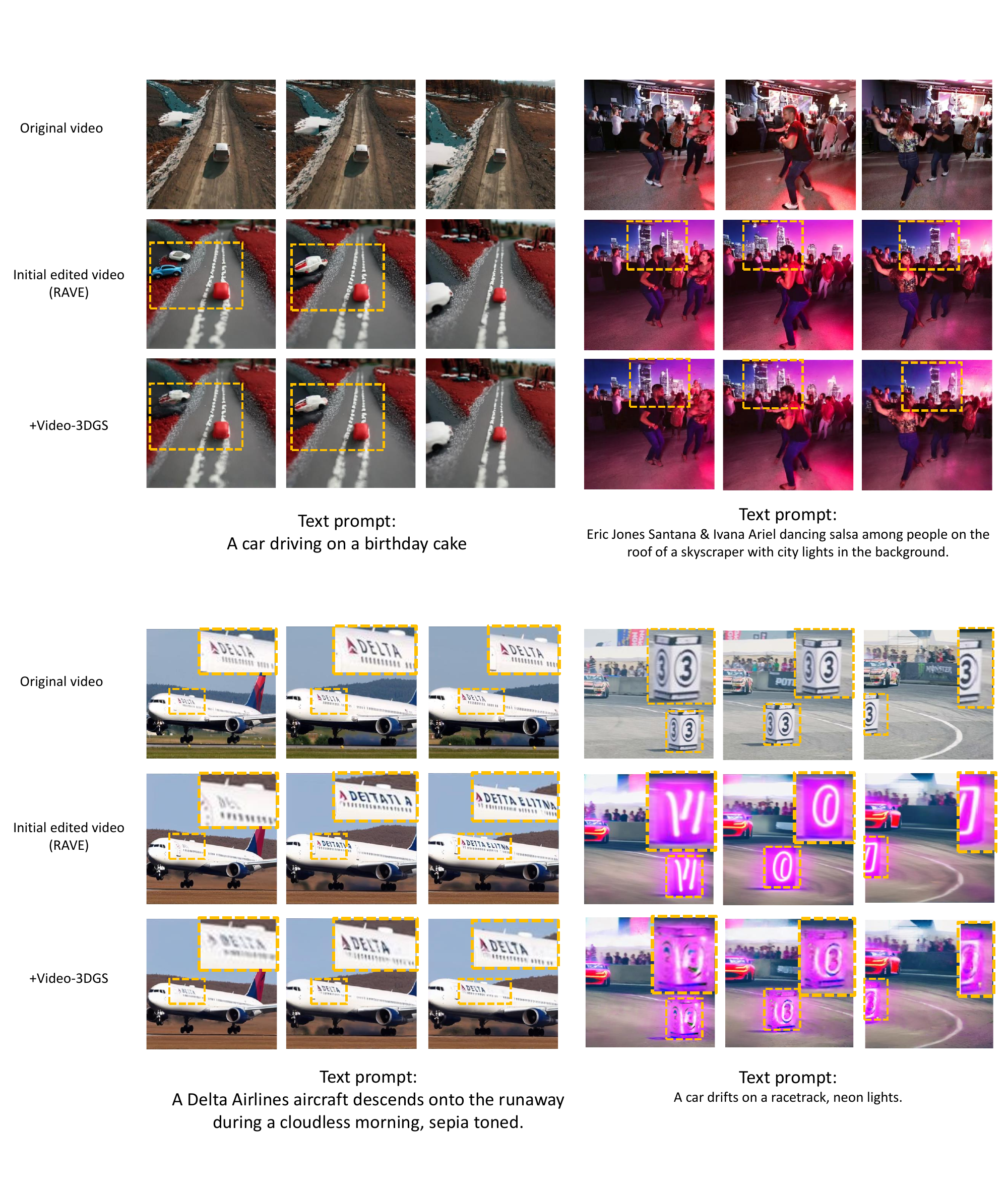}
\vspace{-4mm}
\caption{Qualitative results of single stage refiner for video editing by adopting RAVE.} 
\label{fig:edit3}
\end{center}
\end{figure*}

To show the visual effectiveness of \modelname as single-phase refiner, we visualize qualitative results in~\cref{fig:edit1},~\cref{fig:edit2}, and~\cref{fig:edit3}, which adopt Text2Video-Zero~\citep{text2video-zero}, TokenFlow~\citep{tokenflow2023}, and RAVE~\citep{2312.04524}  as the underlying video editor, respectively. As shown in the figures, \modelname effectively enhances the temporal consistency in the edited results across all three video editors. We further provide the video analysis of \modelname as a plug-and-play refiner to show better visual comparison in this \href{https://anonymous-video-3dgs-additional-edit.github.io/}{link}.

\subsubsection{Recursive and Ensembled Refiner}
We further conduct the visual analysis on \modelname as Recursive and Ensembled (RE) refiner. First, similar to \cref{fig:revisiting}, we provide more video results showing the sensitivity of video editors to hyperparameter changes in this \href{https://anonymous-video-3dgs-re-hyperparameters.github.io/}{link}. It confirms that different hyperparameter values in the deployed video editors results in different outputs. Then, we show the editing visual improvement brought by \modelname (RE) in \href{https://anonymous-video-3dgs-re.github.io/}{here}.

\section{Asset Licenses}
\label{sec:license}
The licenses of the assets used in the experiments are denoted as follows:

Datasets:
\begin{itemize}\small\itemsep0.5em
  \item DAVIS2017~\citep{Pont-Tuset_arXiv_2017}: \url{https://davischallenge.org/index.html}
  \item LOVEU-TGVE-2023~\citep{wu2023cvpr}: \url{https://github.com/showlab/loveu-tgve-2023} 
\end{itemize}

Codes:
\begin{itemize}\small\itemsep0.5em
  \item 3D Gaussian Splatting~\citep{kerbl20233d}: \url{https://github.com/graphdeco-inria/gaussian-splatting}
  \item Text2Video-Zero~\citep{text2video-zero}: \url{https://github.com/Picsart-AI-Research/Text2Video-Zero}
  \item TokenFlow~\citep{tokenflow2023}: \url{https://github.com/omerbt/TokenFlow} 
  \item RAVE~\citep{2312.04524}: \url{https://github.com/RehgLab/RAVE} 
\end{itemize}

Models:
\begin{itemize}\small\itemsep0.5em
  \item Stable Diffusion v 1.5 Models~\citep{rombach2022high}: \url{https://huggingface.co/runwayml/stable-diffusion-v1-5}
\end{itemize}

\section{Broader Impact}
\label{sec:broader_impact}
This paper presents \modelname, which leverages 3D Gaussian Splatting to reconstruct and edit dynamic monocular videos. 
We anticipate significant positive societal impacts from our method, as it consistently performs well in representing diverse videos and refining outputs from zero-shot video editors. We expect it to benefit various applications, such as Entertainment with video synthesis and AR/VR.

\end{document}